\documentclass{article} %
\usepackage{iclr2027_conference,times}

\usepackage{amsmath,amsfonts,bm}

\def\eqref#1{equation~\ref{#1}}

\def\1{\bm{1}}

\DeclareMathAlphabet{\mathsfit}{\encodingdefault}{\sfdefault}{m}{sl}
\SetMathAlphabet{\mathsfit}{bold}{\encodingdefault}{\sfdefault}{bx}{n}

\usepackage{hyperref}
\usepackage{url}
\usepackage{xcolor}
\usepackage{graphicx}
\usepackage{wrapfig}
\usepackage{cleveref}
\usepackage{booktabs}
\usepackage{marginalia}

\usepackage[textsize=tiny]{todonotes}
\newcommand{\ours}{\textsc{Leaps}}
\newcommand{\oursfull}{\textsc{\textbf{Leaps}} (Long-context Evaluation of Adaptive Pattern learning and State tracking)}
\newcommand{\K}{\ensuremath{\text{K}}}

\definecolor{pastelorange}{RGB}{255,224,190}
\newcommand{\hlline}[1]{\leavevmode\kern-1pt{\setlength{\fboxsep}{0pt}\colorbox{pastelorange}{\kern1pt\strut #1\kern1pt}}}

\title{LEAPS: Long-context Evaluation of Adaptive Pattern Learning and State Tracking}

\author{
Howard Yen$^{\hspace{.1em}{\color{orange}\boldsymbol{p}}}$\thanks{Work done as an intern at Jane Street.}$\;$ ~~
Nitya Mani$^{\hspace{.1em}{\color{blue}\boldsymbol{j}}}$ $\;$
Alex Renda$^{\hspace{.1em}{\color{blue}\boldsymbol{j}}}$ $\;$
Christopher S. Hardin$^{\hspace{.1em}{\color{blue}\boldsymbol{j}}}$ $\;$\\
$^{\hspace{.1em}{\color{orange}\boldsymbol{p}}}$Princeton Language and Intelligence, Princeton University $^{\hspace{.1em}{\color{blue}\boldsymbol{j}}}$Jane Street\\
\texttt{hyen@cs.princeton.edu}
}

\iclrfinalcopy %
\begin{document}

\maketitle

\begin{abstract}
   
Long-context language models (LCLMs) power agents that act over long trajectories, from improving a codebase to running research experiments to forecasters that condition on lengthy histories.
These applications draw on three distinct long-context capabilities: \textit{recall}, where the answer is a fact present in the context; \textit{state tracking}, where the outcome can be computed by performing a series of transformations to a specified state; and \textit{pattern learning}, where the goal is to make a prediction by sampling from a distribution inferred from many observations.
Existing benchmarks primarily focus on the first capability.
In this work, we introduce \oursfull, a collection of 12 novel datasets, spanning agentic workflows, behavioral modeling, and forecasting of market and world events.
By construction, \ours{} is resistant to data contamination, extends to millions of input tokens, and uses reliable metrics.
We evaluate 30 API and open models from 8K to 512K input tokens.
\ours{} distinguishes models that perform near perfectly on recall tasks, finding that such models' state tracking and pattern learning abilities can vary significantly and degrade much more dramatically with increasing input lengths---even for pattern learning, where more observations should make learning a theoretically easier task.
Moreover, thinking models often underperform their base counterparts, exhausting their generation budget or oversimplifying the task.
We hope \ours{} can guide the development of the next generation of LCLMs.\footnote{Code and data at \url{https://github.com/howard-yen/LEAPS}}

\end{abstract}

\section{Introduction}

Today, long-context language models (LCLMs) power personal assistants, coding agents, continual learning, and a wide variety of other tasks. 
Such applications frequently require comprehending large amounts of context \citep[][]{kwa2025measuring,yang2024sweagent,wang2025openhands} specific to the task and rely on the underlying model having multiple distinct \textit{long-context capabilities}. %

For example, in order for a coding agent to successfully autonomously edit files and run commands over a long trajectory, it must be able to perform several sub-tasks. 
It should be able to locate code relevant to the current step, which relies on a capability commonly called \emph{recall} \citep{arora2024zoology,hsieh2024ruler}.
It must also be aware of the current state of the repository and its progress on each objective after many edits, requiring strong \emph{state tracking} and understanding \citep{liu2023transformers,merrill2023parallelism}.
Finally, its actions in unseen situations must align with the user's preferences, which it must infer from their distribution of past interactions.
We call this type of in-context learning ability \emph{pattern learning} \citep{brown2020language,garg2022what}.
Many long-horizon tasks, particularly harder ones, frequently make use of all three abilities, which fundamentally differ in \emph{how} models use their available context: to recall facts, maintain and update a relatively comprehensive (but compact) state as changes are applied, or continually learn patterns from observations.

However, existing LCLM development still overwhelmingly relies on recall benchmarks, such as Needle-In-A-Haystack \citep{gkamradt_llmtest_needleinahaystack_2023} and RULER \citep{hsieh2024ruler}.
Some recent benchmarks that evaluate performance on a broader range of realistic applications, such as question-answering, summarization, and reasoning over code \citep[][\textit{inter alia}]{zhang-etal-2024-bench,yen2025helmet,bai-etal-2025-longbench}, many-shot in-context learning \citep{agarwal2024manyshot,bertsch-etal-2025-context,li2025longcontext}, and aggregation \citep{bertsch2025oolong} do broaden the coverage of long-context evaluation, but do not comprehensively study how a model's relative long-context capabilities are apportioned between recall, state tracking, and pattern learning. 

To fill this gap, we systematically study two long-context capabilities---state tracking and pattern learning---by creating a novel benchmark, \oursfull. %
We construct 6 synthetic and 6 real-derived datasets in which LCLMs track ML experiments, imitate top chess players' styles, and forecast events from historical data.
By design, our datasets are robust to contamination (these datasets can be regenerated or sampled past a model's knowledge cutoff), scale to millions of input tokens, and use reliable evaluation metrics.

Our experiments cover 30 API and open models, evaluated from 8K to 512K input tokens. Even models that look very similar on recall tasks can be starkly distinguished on \ours. 
We attribute this in part to \ours\,tasks testing for distinct long-context capabilities and to performance with respect to such capabilities still being relatively unsaturated, even amongst very strong API models.

In contrast to many recall benchmarks, where model performance is relatively consistent across context lengths, performance on state-tracking tasks degrades significantly at long context. Performance on pattern-learning tasks---which become higher signal-to-noise with context length---also declines at long contexts (although sometimes performance can improve at moderate context).%

Studying the correlation of performance in the \ours{} suite across models and context lengths suggests that our datasets evaluate multiple distinct long-context capabilities, not just performance along a single hidden axis. Further, when comparing models, a somewhat surprising observation is that thinking models often perform worse on \ours{} tasks than their base counterparts. Our analysis suggests that they often run out of generation budget without making appreciable progress and can make overly simplifying assumptions, suggesting possible improvements in long-context post-training techniques.
We hope \ours{} can help develop LCLMs for further long-horizon applications.

\section{Our Benchmark: \ours} 

\begin{figure}
    \centering
    \includegraphics[width=0.99\linewidth]{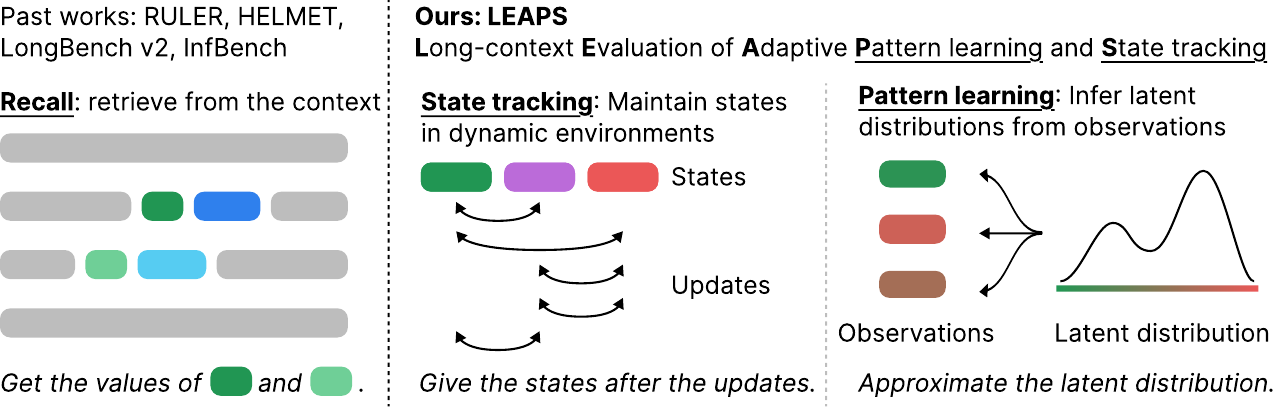}
    \caption{
        In contrast to many previous long-context benchmarks that revolve around \textit{recall}, we study two new capabilities in a long-context setting, \textit{state tracking} and \textit{pattern learning}.
    }
    \label{fig:teaser}
\end{figure}

Many applications require the model to reason over complex contextual information.
In agentic settings, models need to keep track of the goals and the state of the environment over long trajectories---a form of \textit{state tracking}.
In predictive settings, models need to leverage historical observations to approximate latent distributions or user behaviors---an instance of \textit{pattern learning}.
Many long-horizon tasks require a combination of these long-context capabilities, such as a coding agent that has to interact with a file system and the shell while aligning its actions with the user's preferences.

Previous benchmarks often revolve around recall \citep{gkamradt_llmtest_needleinahaystack_2023,hsieh2024ruler,bai-etal-2025-longbench,yen2025helmet}, but state tracking and pattern learning are conceptually distinct because the answer is not present in the context.
Instead, models must compute the answer by using the entire context, either composing state-modifying transformations or inferring patterns from historical data (\Cref{fig:teaser}).
Interestingly, as input length increases, state-tracking tasks remain perfectly solvable and require a constant-sized memory that does not depend on the trajectory length if done efficiently, whereas the ceiling of performance on pattern learning tasks increases as more data is available.

We introduce \ours, a suite of benchmarks which consists of six state-tracking and six pattern-learning tasks, including a mix of synthetic tasks, designed to isolate a narrow capability, and application-centric tasks highlighting a model's performance in downstream settings. 
We design the tasks with the following desiderata:
(1) robustness to data contamination by generating data after the model knowledge cutoff,
(2) scalability to millions of tokens for comparisons across lengths, 
(3) reliable and interpretable metrics, standardized between 0 and 1 with higher being better, %
(4) naive baselines for contextualizing model performances,\footnote{We notice that weaker models may underperform basic strategies, such as guessing a random option, so we provide these baselines as an anchor point.}
and (5) cheap inference costs.
An overview of the datasets is in \Cref{tab:descriptions} and more details for all datasets are in \S\ref{app:dataset-construct}.

\begin{table}[t!]
  \centering
  \small

  \begin{tabular}{@{}llc@{}}
  \toprule
  \textbf{State tracking} & Description & Source \\ 
  \midrule
  Cup swapping & Track positions of balls as cups are swapped. & Syn \\
  Coins & Track the count of piles as coins are added, removed, and swapped. & Syn \\
  Orderbook & Track buy and sell orders in a multi-actor market. & Syn \\
  ML Dev & Follow parallel model developments and generate subsequent commands. & Syn \\
  Commit re-ordering & Identify pairs of conflicting commits. & Real \\
  Code tracking & Track code files across commits and changes. & Real \\
  \midrule
  \textbf{Pattern learning} & Description & Source \\
  \midrule
  Rock-paper-scissors & Learn behaviors of two players following specific strategies. & Syn \\
  Review ranking & Learn preferences of an area chair ranking papers based on reviews. & Syn \\
  Kaggle & Learn relationships between abstract features and market movements. & Real \\
  Social media & Learn relationships between textual features and market movements. & Real \\
  Chess & Learn behaviors of top chess players and predict moves at new states. & Real \\
  Forecast & Forecast real-world events using news articles and past events. & Real \\ \bottomrule
  \end{tabular}

  \caption{Description and data source of each dataset (synthetic or real).}
  \vspace{-1em}
  \label{tab:descriptions}
\end{table}

\subsection{State tracking}

Interactive environments can be formulated as semiautomata,
consisting of a set of finite states $Q$, such as the positions of balls or the content of a file, that are modified by a set of transformations $M$, such as swapping two cups or editing a line of code.
We define state tracking as the ability to compute the state $q_T$ after transformations $\mathcal{M} = \{m_1, \ldots, m_T\}$ on the initial state $q_0$.
Recent works have shown limitations
of existing models on state tracking \citep{merrill2023parallelism,merrill-2024-illusion,liu2023transformers}, and we extend the evaluation to diverse long-context settings.

Individual steps are simple and the number of bits needed to represent $q_i$ remains bounded.\footnote{Code repositories can be arbitrarily long, but we limit the number of files and lines in each task.}
The task becomes increasingly more difficult as more transformations are composed together, because missing just one update can derail the entire computation, but a human can still perfectly solve all tasks with enough time.
Thus, we test how efficiently models can represent and answer questions about $q_T$ as $T$ increases.
We briefly describe each task below and details are in \S\ref{app:dataset-construct}.

\paragraph{Cup swapping.}
In this classic game, balls are placed under exactly one cup, and the task is to maintain the position of the balls as the cups are swapped around.
The input length naturally scales with more operations, and synthetic generation means the model cannot rely on its parametric knowledge for getting the correct final positions.
The metric is the F1 score.

\paragraph{Coins.}
In this task, we test both swapping and linear aggregation operations.
Across many piles of coins, the model must keep track of the count of coins in each pile as coins are added, removed, and swapped between piles. 
Similarly, the input length scales with more operations, and the data is synthetically generated.
The metric is one minus the normalized L1 distance.

\paragraph{Orderbook.}
We test if models can track and reason about states in a market.
We simulate multiple actors sending buy, sell, and cancel orders to a central limit order book, and trades occur as orders arrive.\footnote{The orderbook can grow arbitrarily large, but most orders get canceled or filled by construction.}
The task is to accurately calculate how many shares of a new order would trade and at what average price, which requires a precise account of the book.
The horizon scales with the number of orders, and the data is synthetically generated.
The metric is one minus the L1 distance.

\paragraph{ML Model Development.}
Long-horizon tasks involve progressing on parallel goals while following user instructions. 
We simulate this in an ML model development process with a synthetic shell environment.
The prompt includes the specifications and goals, environment instructions, and a history of executed commands.
The task is producing the next command that advances a specific goal, and the metric is exact match with any valid and useful command; details are in \S\ref{app:dataset-construct}.

\paragraph{Code Tracking.}
Coding agents must keep track of how the code in a repository changes over time to effectively accomplish user tasks.
Thus, we simulate agentic trajectories with commits from GitHub repositories.\footnote{We use repositories with active changes in 2026 and permissive licensing, listed in \Cref{tab:repos}.}
During evaluation, we show the model the initial state of files and subsequent commits and ask the model to output 3--5 lines in the final state of the code.
The metric is exact match, and the length scales with the number of commits.

\paragraph{Commit Re-ordering.}
Coding agents are also commonly used to detect bugs and conflicts.
Using the real-world commits from Code Tracking, we construct sequences of commits in which one pair of consecutive commits is out of order.
The two commits modify overlapping lines, resulting in conflicts if applied out of order, and the model has to identify this pair.
The metric is exact match.

\subsection{Pattern learning}

Real-world events often generate concrete observations from latent distributions $y \sim p_\theta(\cdot \mid x)$, such as a player's chess move at a board state or market movements in response to a piece of news.
We define pattern learning as the ability to approximate $p_\theta$ from observations $\{(x_i, y_i)\}_{i=1}^n$ by predicting $y'$ for a new $x'$.
Pattern learning powers numerous applications, such as forecasting future events based on recent data \citep{halawi2024approaching,karger2025forecastbench}, continual learning in personal assistants \citep{zhao2025do,zhang2026agentic}, and making decisions based on user preferences in unseen situations  
\citep{park2023generativeagents,wang-etal-2023-humanoid} by using numerous observations, making it a key long-context capability to study.

In-context learning tasks are similar in that they also learn from demonstrations $(x,y)$ \citep{brown2020language,xie2022an}; however, they are limited to classification tasks and learning the output format \citep{min-etal-2022-rethinking,bertsch-etal-2025-context,li2025longcontext,yen2025helmet}.
\ours{} is unique in that the observations $(x, y)$ and latent distributions $p_\theta$ are \textit{specific} to each test instance, whereas previous benchmarks draw $(x,y)$ from a pool of shared demonstrations.
We consider diverse latent distributions, spanning player strategies to real market movements. 
Notably, pattern learning is distinct from recall and state tracking in that, holding context degradation aside, performance of strong LCLMs should \textit{improve} with longer inputs, as there are more data to learn from.

\paragraph{Rock-paper-scissors (RPS).}
Humans often exhibit predictable strategies in RPS \citep{dyson2016negative,zhang2021rock,wang2014social}.
We generate synthetic players with random parameterizations that dictate the distribution over their moves, $p_\theta$.
The task is to approximate $p_\theta$ given a history of games, which scales with the input length.
The metric is 1 minus the Hellinger distance.

\paragraph{Chess.}
We study chess as another environment where players have implicit preferences due to their style of play, and we use real games from top players on online websites.\footnote{The games were played from January through May 2026 on \url{www.lichess.org}, a platform popular among top chess players.}
The evaluation input consists of a long history of complete games from a player $p$ and one incomplete game at a board state that $p$ has never encountered before.
We ask the LLM to output what move $p$ will play next, and report the exact match score as the metric.

\paragraph{Review ranking.}
Human preferences result in comparisons and rankings of different items, and the ability to collate preference information into relative rankings is a critical skill in information retrieval \citep{bajaj2016ms,nogueira2020passagererankingbert,sun-etal-2023-chatgpt}. 
To this end, we design a review ranking dataset by generating ``area chairs'' (AC) with random preferences over five axes: Novelty, Significance, Soundness, Presentation, and Reproducibility.
The input includes examples of the AC's ranking over papers, each with multiple synthetically generated reviews, and a set of new papers that the model has to rank. %
The metric is NDCG@20.

\paragraph{Kaggle.}
Forecasting requires inferring relationships between variables, such as how overall market movements affect a specific stock.
We use a popular time-series market forecasting Kaggle competition \citep{jane-street-real-time-market-data-forecasting} that requires using abstract features to predict ticker movements.
We formulate this as a multiple-choice QA task with five approximately equally weighted bins for ease of evaluation.
The metric is exact match (EM).

\paragraph{Social media market forecasting.}
Social media posts reflect public sentiment toward financial markets and can indicate future movements.
We use an internal collection of social media posts, each matched to a specific tradable asset, to evaluate whether models can learn relationships between textual features and market movements.
The prompt consists of posts and subsequent market movements, and the task is to forecast movements on a new set of posts.
Similarly to Kaggle, we construct the task as a multiple-choice QA task and use the same EM metric.

\paragraph{General event forecasting.}
In addition to the stock market, we also consider a wider range of forecasting tasks, such as world events and macroeconomic trends. 
We use the May 2026 questions from  ForecastBench \citep{karger2025forecastbench} and pair each question with relevant news articles from Common Crawl.
The prompt includes examples from an earlier date, each with news articles, a query, and a resolution, followed by the new question and articles. 
The model is asked to output the predicted probability that the query will resolve to true.
We report one minus the Brier score.

\section{Results}

Our evaluations span 30 models from 8K to 512K input tokens.
Across context lengths, \ours{} distinguishes models that appear similar on recall benchmarks by testing for distinct long-context capabilities, and performance on them is still unsaturated.
Even the strongest models fall short of the ceiling on state tracking, and open models often cannot learn from observations effectively in pattern learning.
As the context length increases, existing models experience statistically significant performance degradation on state tracking and pattern learning.
Finally, post-trained models often underperform the base models they are trained from, suggesting that current post-training techniques, which target improving reasoning abilities, can compromise other long-context capabilities.

\subsection{Experiment settings}

We evaluate frontier API models, GPT 5.6 Sol and Terra \citep{openai2026gpt56}, Claude Opus 4.8 and Sonnet 5 \citep{anthropic2026opus48,anthropic2026sonnet5}, and Gemini 3.5 Flash \citep{googledeepmind2026gemini35flash}, and open-weight model families, Llama 3.1 \citep{grattafiori2024llama3herdmodels}, ProLong \citep{gao-etal-2025-train}, GPT-OSS \citep{openai2025gptoss120bgptoss20bmodel}, Qwen 3 and 3.5 \citep{yang2025qwen3technicalreport,qwen35blog}, Gemma 4 \citep{gemmateam2026gemma4technicalreport}, Nemotron Nano \citep{nvidia2025nemotron3nanoopen}, and Kimi Linear \citep{kimiteam2025kimilinearexpressiveefficient}.

Our evaluation covers input lengths from 8K up to 512K tokens, measured in the Llama 3 tokenizer \citep{grattafiori2024llama3herdmodels}.\footnote{Inputs with 512K tokens often translate to more than 600K tokens for many API models. We also release data at 1M tokens but did not evaluate them to leave sufficient context window for generating reasoning traces.}
For base models, we ensure the generation budget is sufficiently large to output the correct answer in each dataset.
For instruction-tuned models, we increase the generation budget by 256 tokens to account for assistant messages.
For thinking models, we increase the budget by 16K tokens to allow long thinking traces, and instruct the model to finish the chain-of-thought within the allotted budget in the system prompt.
We instruct models not to rely on code to solve state tracking to better test if models can do this task with their native representations, which is required in many less-structured real-world applications.
Details are in \S\ref{app:eval-details}.

\begin{figure}[t!]
    \centering
    \includegraphics[width=0.99\linewidth]{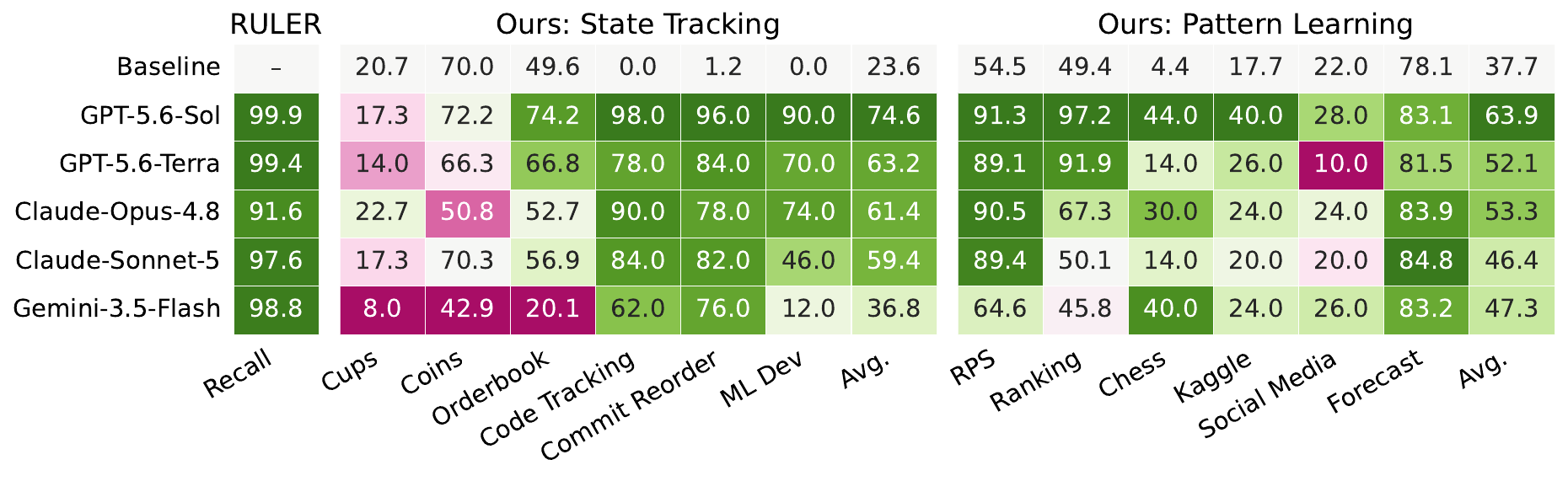}
    \vspace{-1em}
    \caption{
        Results of API models at input length 512K in percentages. 
        Compared with recall tasks, consisting of subsets of RULER and HELMET, \ours{} distinguishes models on two new axes.
    }
    \label{fig:results_512k_api}
\end{figure}

\subsection{\ours{} evaluates distinct long-context capabilities}

\paragraph{\ours{} distinguishes frontier models.}
We present the results of frontier API models in \Cref{fig:results_512k_api}.
The input lengths are 512K tokens, and we use medium reasoning effort for all models, when applicable.
We evaluate on 50 samples from each dataset.\footnote{We discard the samples where the API refuses to respond, which happens less than 4\% of the time.}
Although API models have roughly saturated recall tasks, their performance varies substantially on \ours, allowing for a more nuanced comparison of their abilities on these two long-context capabilities. 
Our findings are complementary to results on other benchmarks, including GraphWalks, MRCR, and LongBench v2 \citep{openai2025gpt41,vodrahalli2024michelangelolongcontextevaluations,bai-etal-2025-longbench} (\S\ref{app:more-results}).

Somewhat surprisingly, these frontier models are often comparable to or even worse than naive baselines on Cups, Coins, and Orderbook. 
We perform additional ablations on these datasets in \S\ref{app:dataset_variants}.
Pattern learning also exhibits similar difficulty---many models barely outperform the naive baselines, especially on the forecasting-related datasets.
Thus, these models still have room to improve.

\begin{figure}[t!]
    \centering
    \includegraphics[width=0.99\linewidth]{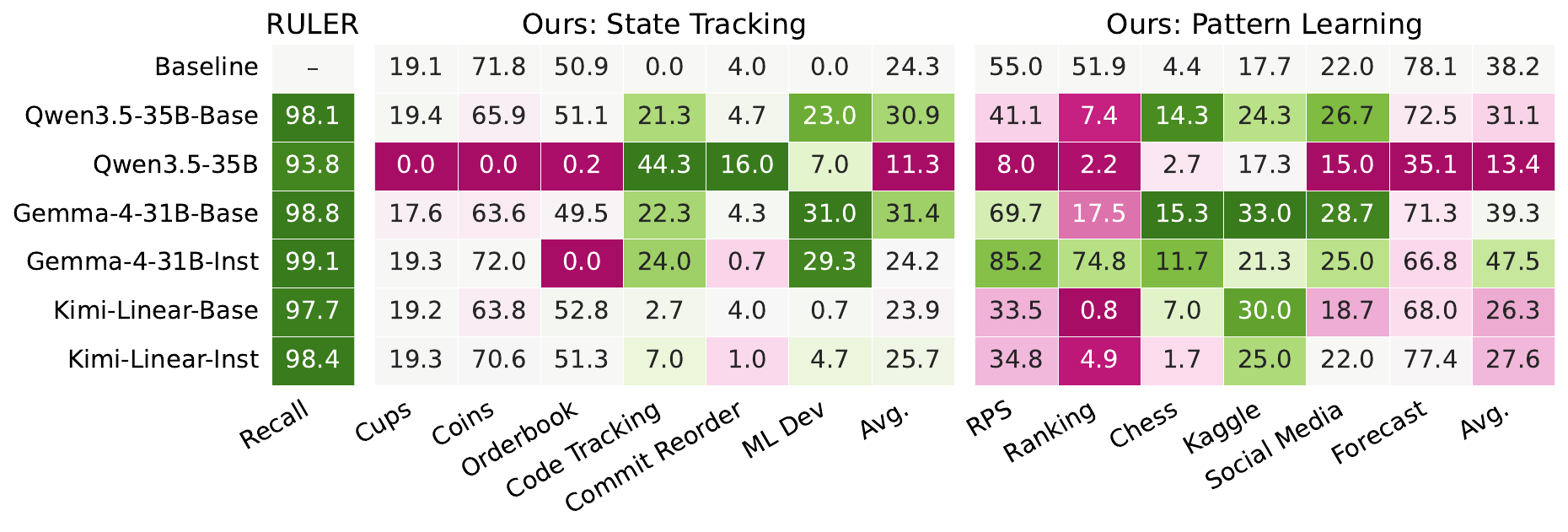}
    \vspace{-1em}
    \caption{
        Results of selected open-weight models at 128K input length. 
        Even the strongest open models struggle with both state tracking and pattern learning.
    }
    \label{fig:results_128k_only}
\end{figure}

\paragraph{\ours{} reveals orthogonal shortcomings of open-weight models.}
We show the results of the high-performing open-weight models at 128K input length in \Cref{fig:results_128k_only}; full results are in \Cref{fig:results_128k_full}. 
Although open-weight models are above 90\% on recall tasks, they exhibit performance similar to or worse than naive baselines on \ours, which suggests that models are still not able to effectively track states across long trajectories or learn patterns from many observations.
Similar to API models, \ours{} highlights two new dimensions of long-context capabilities in open models.

\subsection{\ours{} shows significant degradation with longer length}

\begin{figure}[t!]
    \centering
    \includegraphics[width=0.99\linewidth]{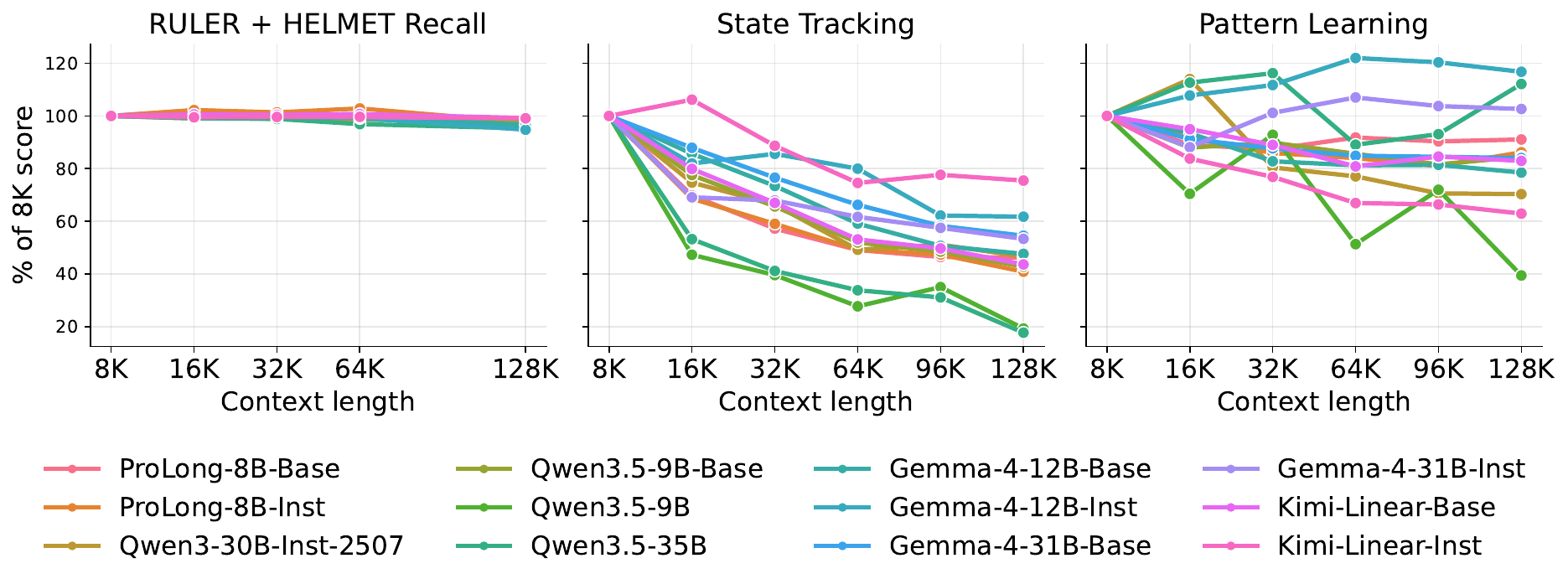}
    \vspace{-1em}
    \caption{
        Performance is shown relative to each model's performance at 8K input tokens.
        Models observe substantially more degradation with increasing length on state tracking and pattern learning compared to recall.
        Model performance drops by up to 80\% from 8K to 128K on \ours.
    }
    \label{fig:results_128k_comp}
\end{figure}

We present the results from 8K to 128K input length for selected models in \Cref{fig:results_128k_comp}, relative to each model's performance at 8K.
The full results and absolute numbers are in \Cref{fig:results_128k_full}.

\paragraph{Variable degradation at long context.} 
The three capabilities show distinct trends as the context plays fundamentally different roles in each: with longer inputs, recall results are relatively consistent, state tracking smoothly declines, whereas pattern learning observes some increase in performance, as strong models can make use of the additional data, before decreasing.

Notably, model performance may drop by as much as 80\% on \ours.
The decrease from 8K to 128K over all models observes greater statistical significance on state tracking and pattern learning than on recall.
State tracking is especially difficult because it requires composing a long chain of transformations, and missing any step would sabotage the performance.
Although few models show improvements on pattern learning with longer inputs, most still fail to leverage more observations.

\begin{figure}[t!]
    \centering
    \includegraphics[width=0.99\linewidth]{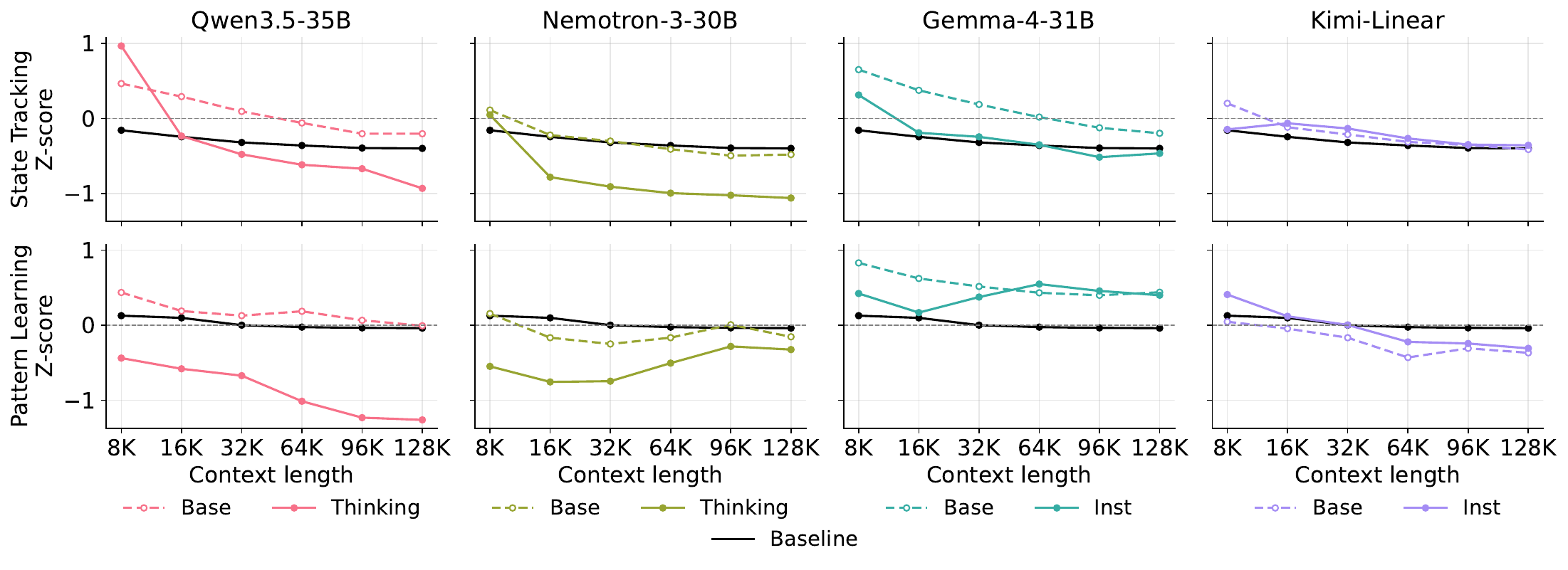}
    \vspace{-1em}
    \caption{
        Comparison of base and post-trained models.
        Results shown in z-scores to standardize averaging across datasets and comparisons across input lengths.
    }
    \vspace{-1em}
    \label{fig:results_base_thinking_comp}
\end{figure}

\paragraph{Thinking models underperform their base counterparts.}

Across different model families of Qwen3.5, Nemotron, Gemma, and Kimi Linear, we find that the base models generally perform better than or equal to their post-trained counterparts in \Cref{fig:results_base_thinking_comp}. 
To fairly assess overall performance across datasets and input lengths, we normalize the absolute performance into z-scores in each dataset before taking the average; details are in \S\ref{app:eval-details}.

Thinking models, Qwen3.5-35B and Nemotron-3-30B, are substantially worse than their base models, by up to 1 standard deviation.
Qualitatively, we find that existing thinking models often spend most of their generation budget on reasoning traces or make wrong assumptions; more in \Cref{aly:behavior}.
In contrast, base models generate better answers by relying on their hidden representations.

Instruction-tuning may result in slight degradation in performance at shorter input lengths, but Gemma-4 and Kimi-Linear achieve similar performance at 128K input length.
Thus, how to post-train effective LCLMs to improve state tracking and pattern learning remains an open question.

\section{Analysis}

\subsection{\ours{} datasets span distinct capabilities}

\begin{figure}[t!]
    \centering
    \includegraphics[width=0.99\linewidth]{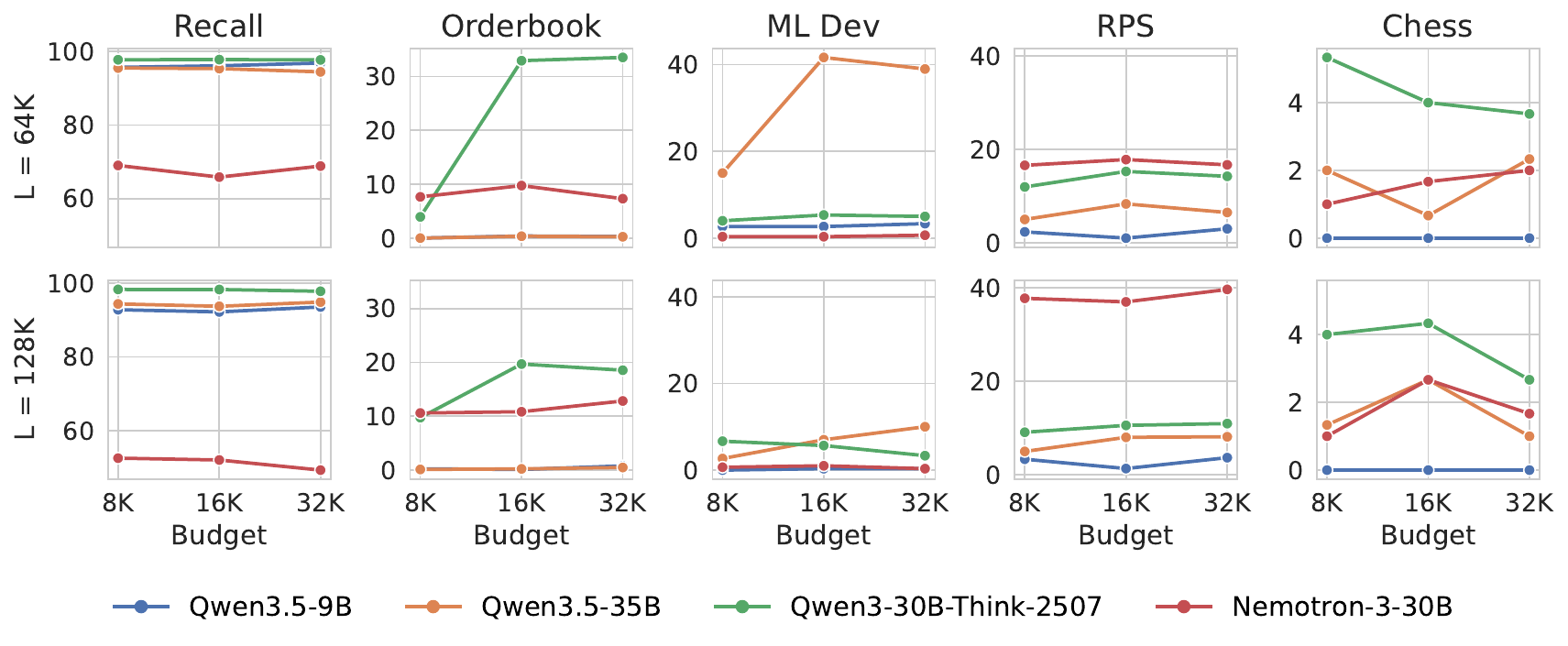}
    \vspace{-1em}
    \caption{
        Results of thinking models with generation budgets from 8K to 32K, on input lengths $L= \{64\K, 128\K\}$.
        Performance may improve going from 8K to 16K output but plateaus after.
    }
    \vspace{-1em}
    \label{fig:analysis_output_length}
\end{figure}

\begin{wrapfigure}{r}{0.46\textwidth}
    \vspace{-2pt}
    \centering
    \includegraphics[width=\linewidth]{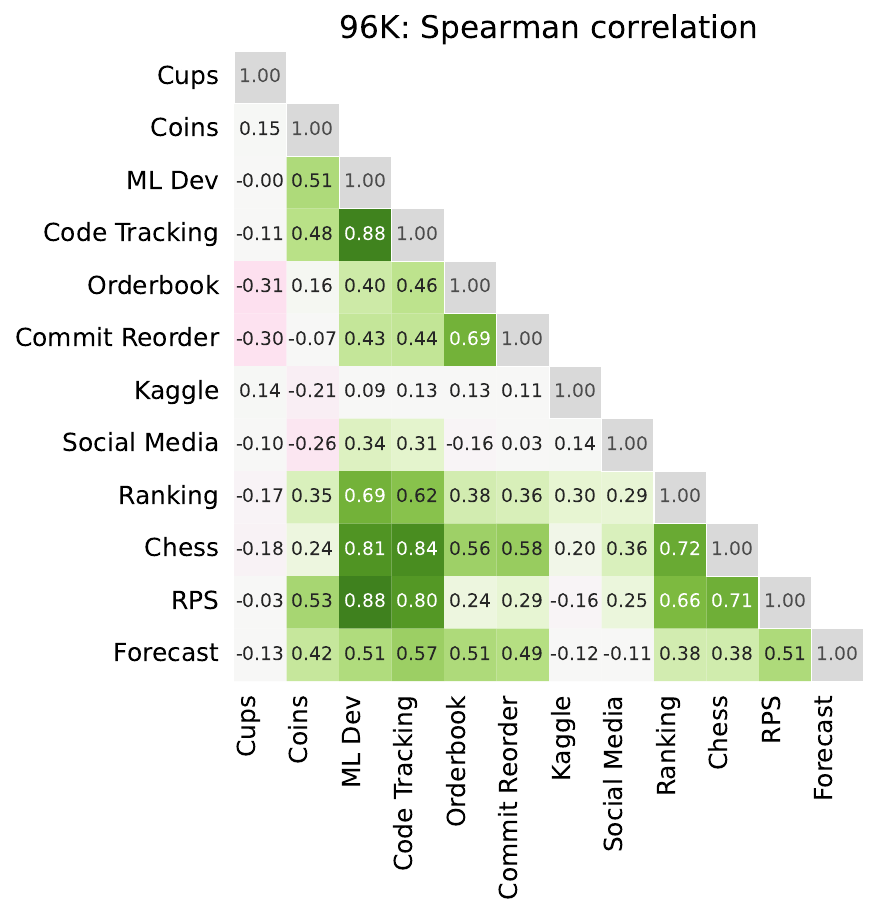}
    \vspace{-15pt}
    \caption{
        Correlation matrix at $L=96\K$, calculated across 23 models: API models and open-weight non-thinking models. The standard error of Spearman $\rho=0$ is $0.209$.
    }
    \label{fig:analysis_corr}
    \vspace{-3em}
\end{wrapfigure}

We plot the correlation between different datasets in \ours{} in \Cref{fig:analysis_corr}.
Intuitively, many of the pattern learning tasks tend to correlate highly with each other---Chess, RPS, and Forecast exhibit significant positive correlation with each other.
However, the market movement prediction tasks, Kaggle and Social Media, exhibit relatively low correlation with all other tasks, suggesting that predicting stock returns may test for different capabilities. 
Thus, the diverse range of datasets allows us to analyze LCLMs along multiple aspects of long-context capabilities.

Furthermore, the state-tracking tasks show moderately strong correlation with each other, with the exception of the Cups dataset that shows low correlation with all other datasets.
Surprisingly, most state-tracking datasets also correlate strongly with pattern learning tasks.
Notably, ML Dev and Code Tracking tasks show medium to strong correlation with Social Media, Forecast, Ranking, Chess, and RPS.
Thus, state tracking and pattern learning capabilities may be intertwined.

\vspace{-1pt}
\subsection{More thinking does not help}

For our main experiments, we give the thinking models a budget of $B=16\K$ extra tokens during generation to try to ensure that generation budget is not a limiting factor; for reference, previous work set $B=8\K$ during long-context evaluation\footnote{Details on \url{https://huggingface.co/Qwen/Qwen3-30B-A3B-Thinking-2507}} \citep{yang2025qwen3technicalreport}. 
However, we observed that many thinking models, especially the Qwen models, do not output a final answer due to running out of generation budget on some tasks in \Cref{fig:analysis_thinking_finish_full}. While reasoning efficiently is \textit{also} an important skill, in these benchmarks, we focused on a model's terminal ability to arrive at the correct answer, and thus ran an ablation where we varied the number of allowed generation tokens between 8K and 32K to test if additional thinking tokens led to better results.
We show the results in \Cref{fig:analysis_output_length}.

Recall observes little change with changing generation budget.
On \ours, performance may increase as $B$ increases from 8K to 16K tokens, suggesting that our tasks are more difficult and require more thinking.
However, performance plateaus past $B=16\K$; some settings may even incur some degradation, such as Chess at 128K. Thus, more thinking is not enough for solving \ours.

\subsection{Model behaviors}
\label{aly:behavior}

\begin{table*}[t]
    \centering
    \scriptsize
    \begin{tabular}{@{}p{0.455\textwidth}@{\hspace{0.05\textwidth}}p{0.455\textwidth}@{}}
        \toprule
        \textbf{Qwen 3.5 30B Cups reasoning} &
        \textbf{GPT-OSS-120B Kaggle reasoning} \\
        \midrule
        The user wants to know the final positions of three balls.

        \(\cdots\)\par

        I need to go through the list of ``On table X, swap cups Y and Z'' instructions and filter for those that apply to Table 58.

        \smallskip
        \hlline{1. On table 58, swap cups 3 and 2.}\par
        \hlline{2. On table 58, swap cups 3 and 4.}\par
        \hlline{\(\cdots\)}\par
        \hlline{816. On table 58, swap cups 5 and 2.}\par
        \hlline{817. On table 58, swap cups}\par
        \textbf{[run out of generation budget]}

        &

        maybe the responder is simply the last value of ``Continuous data''? \(\cdots\)
        Take first example: \(\cdots\) responder equals last continuous data value. 
        Check second example: answer (C) responder 0.0412. Continuous data ends with ``\ldots{} 0.0412''. 
        \hlline{Yes. So responder = last element of continuous data.}

        The continuous data line ends with: ``\ldots{} -0.3101, 2.3888, 1.3782, 0.7434''. 
        \hlline{Yes last value is 0.7434. Thus responder = 0.7434.}

        \smallskip
        Thus output: (E)\par
        \textbf{[Gold answer: (C)]} \\
        \bottomrule
    \end{tabular}
    \caption{Examples of unsuccessful model reasoning at 128K input tokens. Qwen 3.5 30B exhausts its generation budget while explicitly tracking swaps in a Cups sample (left), while GPT-OSS-120B assumes a spurious pattern and returns an incorrect answer on a Kaggle sample (right).}
    \label{tab:example-main}
\end{table*}

Many LCLMs perform similarly to or even worse than naive baselines, which resembles random guessing by minimally using the context; we further investigate the failure cases through qualitative analysis, as shown in \Cref{tab:example-main}. 
We observe two interesting behaviors: relying on brute-force methods in state tracking and making overly simple assumptions in pattern learning.
Many models often manually step through all operations, exhausting their generation budget, suggesting a fundamental limitation of how existing LCLMs use their hidden representations for state tracking.
We also find that existing models can assume spurious patterns in pattern learning, leading to incorrect conclusions. 
For example, GPT-OSS-120B's reasoning chain incorrectly assumes the responder is just the last number in the numerical feature due to two in-context examples, and ignores the rest of the input.
Additional analysis and ablations are in \S\ref{app:dataset_variants}.

\section{Related Works}

\paragraph{Long-context benchmarks.}
Synthetic tasks have been popular in long-context evaluation \citep{hsieh2024ruler,gkamradt_llmtest_needleinahaystack_2023,openai2025gpt41,vodrahalli2024michelangelolongcontextevaluations}.
However, they are limited to retrieval settings, which do not cover other long-context capabilities.
Recent benchmarks have also extended evaluations to question-answering, RAG, summarization, and code analysis \citep{zhang-etal-2024-bench,yen2025helmet,bai-etal-2025-longbench,lu2026corpusqa10milliontoken,lee2024longcontextlanguagemodelssubsume}.
These tasks still primarily test retrieval-related capabilities, whereas state tracking and pattern learning capabilities required by agentic applications and behavioral simulations are still overlooked.

Many-shot in-context learning has also been used to test if models can reliably learn new tasks on the fly \citep{bertsch-etal-2025-context,li2025longcontext,yen2025helmet}, but our pattern-learning tasks extend context learning to more complex settings beyond classification tasks.
CLBench \citep{dou2026clbenchbenchmarkcontextlearning,dou2026clbenchlifelanguagemodels} and LongProc \citep{ye25longproc} also test the ability to learn from context, but they are more instruction-focused and cannot be easily scaled to longer contexts.
Aggregation tasks have also been studied \citep{bertsch2025oolong}, but our state-tracking datasets cover more agentic usage of LCLMs.
State tracking \citep{toshniwal2022chess,liu2023transformers,merrill-2024-illusion} and forecasting \citep{tan2024forecast,gruver2023large,halawi2024approaching} have been studied in short-context settings before, and we are the first to apply them in long-context settings.

\paragraph{Long-context language models.}
LCLMs power many powerful long-horizon agents \citep{yang2024sweagent,wang2025openhands}, but the development of these models still evaluates on few synthetic tasks \citep[][\textit{inter alia}]{glm5team2026glm5vibecodingagentic,gemmateam2026gemma4technicalreport,kimiteam2025kimilinearexpressiveefficient,gao-etal-2025-train}.
Furthermore, many behavioral studies leverage LLMs for social simulations \citep{park2023generativeagents,wang-etal-2023-humanoid,ziems-etal-2024-large}, and LCLMs can be a powerful tool for ingesting and learning user behaviors.
Continual learning has also emerged as an powerful paradigm for adapting LLMs to domain-specific and user-specific applications \citep{zhao2025do,zhang2026agentic}, and LCLMs serves a critical role in ingesting the additional context and updating accordingly.
In this work, we develop \ours{} to further the progress of building more holistic and powerful LCLMs by enabling evaluating the critical long-context capabilities in these settings.

\section{Conclusion}

We introduce \ours{} for evaluating two critical yet understudied long-context capabilities---\textit{state tracking} and \textit{pattern learning}.
\ours{} consists of 6 novel datasets per capability, including agentic coding and forecasting.
Our benchmark includes reliable evaluation metrics, scales to millions of tokens, and is robust to contamination. 
Examining thirty models at context lengths from 8K to 512K tokens, we find that even frontier models struggle much more with state tracking and pattern learning as compared to more classical long-context recall-oriented tasks.
Thus, these benchmarks are better able to distinguish model performance along these two new long-context axes, both of which continue to hold room for improvement, even for the strongest models. 
Finally, our analysis shows that existing models degrade significantly more with longer lengths, and many thinking models struggle on \ours{}.
We hope \ours{} can be a resource for building the next generation of LCLMs.

\subsection*{AI use statement}

In this work, we used generative AI tools to implement code for procedural generation of synthetic data, data/result visualization, and raw data pre-processing.
We have not used generative AI tools for proposing/refining hypotheses, designing/providing feedback on research methodologies/experiments, supporting qualitative and thematic data analysis, or interpreting results.
Developing theoretical models, formulating/proving mathematical claims, writing proofs, and assisting with translation are not applicable to this work.
Additionally, we used generative AI tools for proofreading and grammatical assistance, but the initial draft and most subsequent revisions were entirely done by humans.
We have reviewed all AI-assisted work. Specifically, LLM-generated code was verified by 2 authors and tested for correctness by at least one author.
We take responsibility for the final content of this work,
including text, claims or artifacts produced with the aid of generative AI.

\subsection*{Reproducibility statement}

We release the code and instructions for generating the data and evaluating the results.
We also release the original evaluation data as Parquet files for ease of use upon publication, with the exception of the social media benchmark due to licensing restrictions.
The evaluation and data generation code can be found in the supplemental materials.
The data is excluded from the supplemental materials due to the limited size allowed by OpenReview.
Specific descriptions of the construction of the data and the experimental settings can be found in \S\ref{app:dataset-construct} and \S\ref{app:eval-details}.

\subsubsection*{Acknowledgments}
We would like to thank Ahmed Ahmed, Nicholas Roberts, Lucas Torroba-Hennigen, Abbas Zeitoun, and others at Jane Street for their feedback and assistance with project infrastructure.
We also acknowledge Danqi Chen, Jeff Cheng, Yoonsang Lee, and other members of Princeton Language and Intelligence for their helpful discussions.

\bibliography{custom}
\bibliographystyle{iclr2027_conference}

\appendix

\section{Appendix}

\subsection{Dataset construction details}
\label{app:dataset-construct}

In this section, we describe each dataset in more details.
Generally, we design the datasets with the following desiderata:
\begin{enumerate}
    \item Robustness to data contamination: due to the secrecy behind the training data of most language models (both proprietary and open-weight ones), we rely on synthetic generation or data sources that can be easily updated for our benchmark.
    \item Scalability to millions of tokens: as the context model of modern LLM gets longer, all of our datasets can also scale to millions of tokens. This involves increasing the number of transformations in state tracking or the number of observations in pattern learning.
    \item Reliable and interpretable metrics: useful evaluations stem from having useful metrics that can reliably indicate the outcome with high signal and low noise. In order words, scoring better on our metrics always means the output is better than another output that scores lower, devoid of any randomness or noise that may be present in n-gram based metrics (e.g., ROUGE) or model-based metrics (i.e., due to model stochasticity and bias). 
    \item Grounding baselines: we also use baselines to ground the performance of existing language models, as the baselines can illustrate how difficult a task is expected to me. In general, our baselines are our best estimate on how well a strategy can reasonably perform if it does not carefully examine the context and make its best guess. Some datasets do not have baselines due to the nuances describes in their respective sections.
    \item Cheap inference costs: sampling from models can be extremely expensive, especially with long input lengths; thus, the solutions in our datasets are short, less than 10 tokens in most datasets and almost always under 100 tokens, to ensure low inference costs.
\end{enumerate}

\paragraph{Cup swapping.}
This dataset is based on the classic cup swapping game, where balls are placed under cups, and the task is to keep track of which cup each ball is under.
There are $k$ tables, and each table has $n$ cups and $m$ balls, where $m < n$.
Each ball is placed under exactly one cup, and we simulate random swap operations by randomly sampling a table and a pair of cups in that table to swap.
Humans can often do this task perfectly with minimal assistance; usually just a pen and paper are sufficient for tracking a large number of balls due to the simplicity of the task.  
This task tests the capacity of model state representation. 
In our main setting, we use $k=100$, $n=5$, and $m=3$, and only query for the positions of balls on one table.
That is, we only ask for the positions of the three balls on one particular table.
We sample increasingly more operations to increase the input length. 

For evaluation, we report the F1 score between the generated positions and the gold positions, and the baseline is random positions for the queried balls.
We chose this as the baseline because a person who has lost track of the position can only perform their best guesses by randomly choosing the positions.
In practice, we found that existing models always outputted exactly the queried balls in the output, which means the F1 score is practically recall.

\paragraph{Coins.}
Similar to cups, there are $k$ tables with $n$ piles of coins each.
The key difference is that this task also includes a different type of operation: linear aggregation on each pile.
Specifically, at each time step, we randomly sample an action, which can either swap two piles or add/remove a coin from one pile. 
We limit the number of coins in each pile to between 0 and 100---if a pile already has 100 coins, it cannot be sampled to add another coin.

During development, we had initially set the probability of adding and removing a coin to be equal for all piles.
That is, for any given pile, the probabilities for adding, removing, and swapping are $45\%, 45\%, 10\%$, respectively.
As a result, in expectation, the absolute number of coins in each pile stays the same, and given the relatively low chance of swapping, a very good heuristic was simply guessing the initial state of each pile.
However, this doesn't require the model to perform much state tracking---guessing that there are roughly the same number of adds and removes is not really a long-context capability.
Thus, for each pile, we use a beta distribution $p \sim \mathrm{Beta}(5, 5)$ to determine the probability of adding a coin for that pile (versus removal), which increases the variance of changes in order to thwart the trivial heuristic.

In our final setting, we use $k=100$ and $n=5$, and we only ask about 3 piles in evaluation. 
Given that the number of coins is limited to between 0 and 100, we can normalize the L1 distance of the prediction by dividing by 100. Then, we report the average of the three normalized L1 distances, subtracted from 1, as the metric.
The naive baseline is outputting the initial state of each pile, because one might just guess no change after seeing roughly the same number of additions and subtractions\footnote{Another naive baseline could be guessing 50 (the midpoint) coins in each pile, but the instruction does not say that there can be at most 100 coins in each pile, and it may be an unreasonable guess if there were 1 coin in each pile at the start, for example.}.

\paragraph{Orderbook.}
We simulate a central limit orderbook to test if models can keep track of a highly dynamic environment.
Specifically, we synthetically generate 10 actors who all trade in one market.
Each actor has an internal fair value (also denoted as ``fair''), which is how much the actor values the security and is initialized by a Gaussian with mean 100 (an arbitrary choice).
Then, at each step, we randomly sample an actor to take an action
of buy, sell, or cancel.

When buying or selling, the actor samples random noise to add to its internal fair and to size its order.
After an order is sent, it can result in one or more trades if there are matching resting orders (which cannot be from the same actor by construction).
After a trade, both actors update their internal fairs by decreasing the fair if they just bought or increasing the fair if they just sold. 
After their fair is updated, each actor cancels any existing orders of theirs that would buy at a price higher than their fair or sell at a price below it.

The input context consists of two demonstrations of this task and just the actions; no transactions are shown.
We simulate this until the length has reached the target number of tokens, and pose a question: If a particular new order were to arrive, how many shares would trade, and at what average price?
This requires the model to keep track of existing orders (incorporating the effect of any cancels or trades that have taken place). 

For the metrics, we report 1 minus the average between the normalized L1 distances for both the quantity and the average price.
The quantity is normalized by dividing by the quantity in the query, as the number traded cannot exceed the latter.
The guessed price is normalized by first clamping the value between the query price and the most aggressive price on the other side; for example, for a buy order, we would clamp to the window bounded by the lowest resting sell order price and by the query price.
Then, we divide the L1 distance by the size of this window.
If the number traded is 0, then the model is only rewarded with a score of 1 if both the output query and quantity are 0; otherwise, the score is just 0.
The naive baseline is just the query price and quantity.
If one has lost track of the state, the only anchor point they would have is the query price and quantity.

\paragraph{ML Development.}
We simulate an ML model development environment, where the model has to follow specific instructions and steps to accomplish the model development goals outlined by the user.
Specifically, the user begins with an instruction on how many GPUs to use during training and evaluation, which are sampled randomly, and all subsequent commands must follow that.
Then, the model is shown multiple goals, each of which consists of a model size (6 possibilities), a training corpus (6 possibilities), a grid of learning rates and batch sizes (randomly sampled floats), a list of benchmarks to evaluate on (5 possibilities), and a production target to push to (randomly sampled UUID).
The number of goals and the size of the learning rate and batch size grid increase with the target input length.

Then, to accomplish each goal, the model must issue the following commands, in order with the right arguments: train the model on the correct configuration (each one is run separately), inspect the train loss curves, evaluate on the designated downstream tasks, print a table summarizing the results for a set of models (only valid after all models have been evaluated), and push the best model according to the benchmark to production.
In this environment, we can keep track of exactly all valid commands at any given step, and each command is mapped to exactly one goal that it advances.

The input also includes two demonstrations of example trajectories, both of which are relatively short.
Finally, we show the model a history of the commands that have already been run and ask it to output another command that advances a specific goal.
We report the metric of EM to any of the valid commands, and there is no naive baseline here, since the number of possible commands is infinite (or equivalently, running a random command would have a probability of 0 of being correct).

\paragraph{Chess.}
We first download games from the Lichess database\footnote{\url{https://database.lichess.org/}} from the months of January 2026 to May 2026.
We filter for all players who played at least 100 games and at least one at a rating of 2800 in any mode in May of 2026. We also filter out all bots (tagged by the website, checkable via the provided API call).
Generating new games can be done by downloading the latest games.

We include metadata in the prompt: the date and time, time control, event (if applicable), and username and Elo of each player.
For the moves, we show the progression of each game with the standard chess algebraic notation with the time remaining after each move. 
For each player, we keep track of all chess states they have encountered throughout the first five months of 2026, and only evaluate on the first time they encounter a board state.

In the prompt, we show two demonstrations, each with two games in the history and one game to be formatted in the unfinished state. 
Then, the rest of the context is filled with the consecutive games that the player has played leading up to the test game.
The test game has to last at least 20 turns, and we only ask questions about a turn at or after the 20th turn, so that it is less likely for the move to be part of a standard opening line.
Notably, the board is in a state that the player has not encountered in any of the demonstration games, which means that the model cannot simply retrieve the move that the player has played at that state before.

We report the exact match score with the actual move played as the metric.
The naive baseline is outputting a random legal move at that state.
Although it would have been interesting to explore how the model's moves differ from a chess engine's moves, specific chess analysis is out of the scope for this work.
For example, one might use a chess engine, such as one in the StockFish family, to estimate the move a chess player of a particular elo rating would play, and use that as a baseline to understand how much better a language model is at predicting the move.
We leave such explorations for future work.

\paragraph{Code tracking and commit reordering.}
We source the commits from a list of permissively licensed repos for both code tracking and commit reordering.
We select repos with active commit history in 2026 to avoid contamination with pre-training data of LLMs.
The full list of all the GitHub repos is in \Cref{tab:repos}.
For both tasks, we also put a simple toy example of the task at the beginning of the prompt as an in-context learning demonstration.

\textbf{Code tracking.} We first process all commits in 2026 and track metadata like which files each commit changes and the diffs from each commit. 
Then, we create an anchor for each month from January to May, which is that month's first mainline commit, which will be used as the starting point for all test cases.
We also filter out patches that only edit binaries, lock files, and other non-code or large individual files over 100KB. 

Then, for a particular target input length, we first reserve the first 20\% of the context window for showing the initial state of the files, and the remainder is for the commit messages and diffs.
We pack the remaining context window with consecutive commits from the first day of the month, and then choose the initial files to display based on which ones were modified.

Then, we choose a range of file lines to ask questions about. 
We first look at the git diff between the starting and ending states.
We reject the line ranges where the Markdown fences are included to make parsing later easier.
For each starting anchor, we ask at most five questions about different sections of the files.
We report the exact match between the output and the answer.
The baseline is to guess the original state of the code in the file, but by construction that is 0 because we only query the lines that have been modified.

\textbf{Commit reordering.} 
We don't need to use the anchor at the start of every month as we don't need to show the initial state of each file. 
We build sequences of commits starting from the first day of 2026 and break them into chunks of the target lengths. Windows with fewer than four commits are discarded. We also filter out diff patches on binary files, lock files, and other oversized files.
We render each commit with its commit message and diffs and remove the git blob hashes.
We then find all consecutive pairs of the commit messages with dependencies---pairs where the second commit modifies (or removes) a line that was added by the first commit.
Finally, we randomly swap one such pair and ask the model to identify it.
We report the exact match score as the metric, and the naive baseline is outputting a random pair, which is $\frac{1}{n-1}$ in expectation, where $n$ is the total number of commits.

\begin{table}[htbp]
  \centering
  \small
  \begin{tabular}{lll}
    \toprule
    \textbf{Repo} & \textbf{URL} & \textbf{License} \\
    \midrule
    Fabric               & \url{https://github.com/danielmiessler/Fabric}      & MIT \\
    PowerShell           & \url{https://github.com/PowerShell/PowerShell}      & MIT \\
    Vibe-Trading         & \url{https://github.com/HKUDS/Vibe-Trading}         & MIT \\
    agency-agents        & \url{https://github.com/msitarzewski/agency-agents} & MIT \\
    astropy              & \url{https://github.com/astropy/astropy}            & BSD-3-Clause \\
    biome                & \url{https://github.com/biomejs/biome}              & MIT/Apache-2.0 \\
    codebase-memory-mcp  & \url{https://github.com/DeusData/codebase-memory-mcp} & MIT \\
    curl                 & \url{https://github.com/curl/curl}                  & curl \\
    django               & \url{https://github.com/django/django}              & BSD-3-Clause \\
    flash-attention      & \url{https://github.com/Dao-AILab/flash-attention}  & BSD-3-Clause \\
    ghostty              & \url{https://github.com/ghostty-org/ghostty}        & MIT \\
    gin                  & \url{https://github.com/gin-gonic/gin}              & MIT \\
    json                 & \url{https://github.com/nlohmann/json}              & MIT \\
    llama.cpp            & \url{https://github.com/ggml-org/llama.cpp}         & MIT \\
    maigret              & \url{https://github.com/soxoj/maigret}              & MIT \\
    memos                & \url{https://github.com/usememos/memos}             & MIT \\
    mlx                  & \url{https://github.com/ml-explore/mlx}             & MIT \\
    narwhals             & \url{https://github.com/narwhals-dev/narwhals}      & MIT \\
    no-mistakes          & \url{https://github.com/kunchenguid/no-mistakes}    & MIT \\
    numpy                & \url{https://github.com/numpy/numpy}                & BSD-3-Clause \\
    ollama               & \url{https://github.com/ollama/ollama}              & MIT \\
    openclaw             & \url{https://github.com/openclaw/openclaw}          & MIT \\
    openpilot            & \url{https://github.com/commaai/openpilot}          & MIT \\
    oxidizer             & \url{https://github.com/microsoft/oxidizer}         & MIT \\
    pixi                 & \url{https://github.com/prefix-dev/pixi}            & BSD-3-Clause \\
    polars               & \url{https://github.com/pola-rs/polars}             & MIT \\
    pytest               & \url{https://github.com/pytest-dev/pytest}          & MIT \\
    rattler              & \url{https://github.com/conda/rattler}              & BSD-3-Clause \\
    ripgrep              & \url{https://github.com/BurntSushi/ripgrep}         & Unlicense/MIT \\
    ruff                 & \url{https://github.com/astral-sh/ruff}             & MIT \\
    scikit-learn         & \url{https://github.com/scikit-learn/scikit-learn}  & BSD-3-Clause \\
    semantic-kernel      & \url{https://github.com/microsoft/semantic-kernel}  & MIT \\
    sympy                & \url{https://github.com/sympy/sympy}                & BSD-3-Clause \\
    torchtitan           & \url{https://github.com/pytorch/torchtitan}         & BSD-3-Clause \\
    ty                   & \url{https://github.com/astral-sh/ty}               & MIT \\
    uv                   & \url{https://github.com/astral-sh/uv}               & Apache-2.0/MIT \\
    vision               & \url{https://github.com/pytorch/vision}             & BSD-3-Clause \\
    zstd                 & \url{https://github.com/facebook/zstd}              & BSD-3-Clause \\
    \bottomrule
  \end{tabular}
  \caption{Repositories and licenses used for the code tracking and commit reordering datasets.}
  \label{tab:repos}
\end{table}

\paragraph{Rock-paper-scissors.}
Humans often exhibit predictable strategies in RPS \citep{dyson2016negative,zhang2021rock,wang2014social}.
We simulate players with the strategies below:
\begin{enumerate}
    \item Fixed random: probabilities are randomly sampled and fixed regardless of the opponent's moves. Note that this is not how humans play this game, but we add this for variety.
    \item Win-Stay/Lose-Change (WSLC): a win leads to an increase in the probability of playing the same move by $\Delta p$; a loss can lead to either change 50/50, upgrade (change to the move that would have beaten the move it just played), or downgrade (change to the move that would have lost to the previous move).
    \item Win-Change/Lose-Stay (WCLS): the opposite of WSLC.
    \item Adversarial: given the opponent's most popular move in the past $m$ turns, play the move that beats that move. Evenly split the probability if there are multiple most played moves by the opponent.
\end{enumerate}

Each player's strategy is randomly chosen and their parameters, such as $\Delta p$ and $m$, are also randomly initialized. 
There are an infinite number of possible players and game histories, but we always have a reliable latent model of each player's distribution at any point.
We show the model a history of games played between two players, and we ask the model to output the probability distribution over each player's next moves.

The metric is one minus the average Hellinger distance, and the baseline is always guessing $\frac{1}{3}$ for all moves, as a reasonable prior is that people would have equal preference over three random moves.

\paragraph{Review ranking.}
We simulate an area chair who has to rank a list of papers. 
The area chair has a hidden preference over five axes: Novelty, Significance, Soundness, Presentation, and Reproducibility. 
The hidden preference $\vec{w}$ consists of five uniformly sampled weights between $[0, 1]$.

Each paper $p$ consists of five reviews, each of which can be mapped to a vector of scores $\vec{r}_{p,i}$ over the five axes.
The score for each axis is an integer from 1 to 5, inclusive.
We determine the scores for each review in the generation process, where a hidden Markov model (HMM) generates a set of pre-defined phrases given a list of scores.
That is, each score on one axis maps to a list of eight pre-defined phrases, and the HMM emits one of them at random given a score.
The entire review is a random shuffle order of these phrases.
Please refer to the codebase for the mapping from numerical values to the specific phrases. 

Then, the overall score vector for a paper is simply the average $\vec{r}_p = \frac{1}{n}\sum_{i=1}^n \vec{r}_{p,i}$, and the AC's preference score for the paper is $\vec{r}_p \cdot \vec{w}$.
The AC's preference over the given papers is sorted by their preference scores, and the model needs to learn that preference from the demonstrations and apply it to the new set of test papers.
The number of test papers increases as the input length increases.

The metric is NDCG@20, and the baseline is returning the same order in which the papers appeared.
This should be equal to random guessing in expectation because all papers are generated randomly and the scores have no dependency on the ordering.

\paragraph{Kaggle.}
We use a time-series market forecasting Kaggle competition for testing if models can learn relationships between abstract features and market movements \citep{jane-street-real-time-market-data-forecasting}.\footnote{\url{www.kaggle.com/competitions/jane-street-real-time-market-data-forecasting}}
Both the data and the features are anonymized, which means that there is no identifying information in the text that might link any specific examples to a ticker. Furthermore, the fine granularity of the time interval means that it is highly unlikely for the specific timestamp to have appeared in the training set of any LLMs.
Regardless, it should be relatively easy to re-generate this dataset with fresh data by collecting market data.

In this task, the input is a long history of demonstrations of the tuple features and a responder (i.e., the market movement in the next $x$ minutes). 
The test instance only shows the features, and the model must predict the responder. All demonstrations are the tuples for the same ticker in the timeframe immediately preceding the test instance, and the number of demonstrations is dependent on the target input length.

In our preliminary experiments, we asked the model to directly output the responder, which could be a float with many decimal points, and calculated distance as the metric.
However, this was difficult to interpret, and it was unclear how to normalize the distances between 0 and 1 for the sake of averaging with other datasets.
Furthermore, models often performed pretty poorly when prompted to output a number, especially some weaker models.
Thus, we formulated it as a multiple-choice question-answering task, with five approximately equally weighted bins, such that picking any fixed choice is equal to random guessing in expectation.

The metric is exact match (EM), and the baseline is always picking (C), which corresponds to no movements and is equal to random guessing in expectation.

\paragraph{Social Media Market Forecasting.}

Forecasting is a powerful application of LCLMs, and we also test the ability to predict market movements from textual features using an internal dataset. 
While we cannot release the underlying data, in this task we match content from social media platforms with tradable assets and measure our ability to predict the market movement of an asset in the time interval corresponding to the associated collection of relevant social media posts from that period. 
In our dataset, our test rows are given metadata and textual features, and we fill in context with demonstrations of feature responder pairs from previous timestamps.
Similarly to Kaggle, we also formulate this task as an MCQA task with five bins.
The metric is also EM, and the naive baseline is always choosing (C), which is the same as random guessing in expectation. 
A natural guess for predicting market movements is simply guessing no movement, and this results in approximately 20\% in EM (analogous to random guess).

\paragraph{Forecast.}
We use ForecastBench \citep{karger2025forecastbench} as the source of forecast queries.
These queries span many domains, including prediction markets (e.g., Kalshi, Manifold) and general world events (e.g., Wikipedia, FRED). 
We use the data from the query sets released for May 7th and May 24th of 2026.
These queries are resolved a week later, which are May 14th and May 31st, respectively.
In general, the articles are crawled on the date that they are published, even though CC News doesn't guarantee this.
Refer to the original work for additional details on the data.

The contribution of our work is augmenting each query with news articles from CC News, which consists of a list of crawled articles for every day from curated sources.\footnote{\url{https://commoncrawl.org/blog/news-dataset-available}}
Specifically, we first extract and process all news articles from May 1st to May 24th. 
Note that if an article was crawled on May 24th, it means that it was published then at the latest, which means that it could not have any information on what actually happens later---the answer/resolution cannot exist in any of the articles, and the model must infer the answer from the available information.
We build a retrieval corpus for each day of the news using BM25S \citep{bm25s}, an implementation of BM25 \citep{robertson2009bm25}.
For each query, we retrieve the top $k$ articles from each day. 
Both $k$ and the number of days we include preceding the query date scale with respect to the input length we are targeting.

We use the queries from May 24th as the test instances and the May 7th set as demonstrations for the model to learn from.
We use 1 minus the Brier score as the metric, and the naive baseline is the market value at the freeze date, which comes before the query date, and 50\% if the market value is not available (i.e., the task is from a source that is not a prediction market).

\subsection{Evaluation details}
\label{app:eval-details}

For the API models, we evaluate 50 samples on each dataset due to cost, and for all other models, we evaluate on 300 samples per dataset.
For all evaluations, we use vLLM version 0.23.0 and Transformers version 5.12.1.
The generation token is tuned for each task such that it is always possible for the model to generate the correct answer within the generation budget.
For base models, we directly complete the input prompt and generate the final answer.
For instruction-tuned models, we use model-specific chat templates and add 256 tokens to the generation budget to account for overhead of assistant messages.
For thinking models, we use the model-specific chat template and reasoning parser, and add 16,384 tokens to the generation budget to enable longer chains of thought.
Note that this often implies that many models will degenerate at 128K input length as the total tokens between input and output will exceed their native context window, resulting in near-zero performances. However, we believe that it is still important to stress-test models at long contexts and check whether they degenerate near their context limit.

For all models, we follow their instructions for long-context extension; for example, we apply the YaRN extension settings for the Qwen3 models as specified on their model card.

For \Cref{fig:results_base_thinking_comp}, we show the results in terms of z-scores.
The datasets have different means and standard deviations for their scores, which also span different ranges (e.g., forecasting is almost always between 70 and 80 on our metric, whereas RPS can range from 0 to 100). 
Consequently, an unweighted average would cause the high-variance datasets to dominate the model comparison.
So, we standardize the scores to z-scores before averaging them for the per-capability metric.
Specifically, for each dataset, we take the scores from 8K to 128K across different models, calculate the mean and standard deviation (shown in \Cref{tab:results_mean_std}), and
compute the z-scores $z = \frac{x - \mu}{\sigma}$.

\subsection{Dataset variants}
\label{app:dataset_variants}

For our main experiments, we decided to stick with one headline setting, but our design enables much deeper and more interesting analysis for understanding how models behave in these tasks.
For example, we can experiment with different classes of operations as well as present information in different ways.

We ablate design decisions in the Coins and Orderbook datasets for this analysis.
For Coins, we increase the probability of swapping any two piles from 10\% to 90\%, which means that most of the operations go from linear aggregation (i.e., adding/removing a coin from one pile) to non-commuting operations (i.e., swapping piles).
For Orderbook, we present the actions in reverse chronological order instead of forward order, which enables us to analyze if the order of state operations affects the model performance.

\begin{figure}
    \centering
    \includegraphics[width=0.99\linewidth]{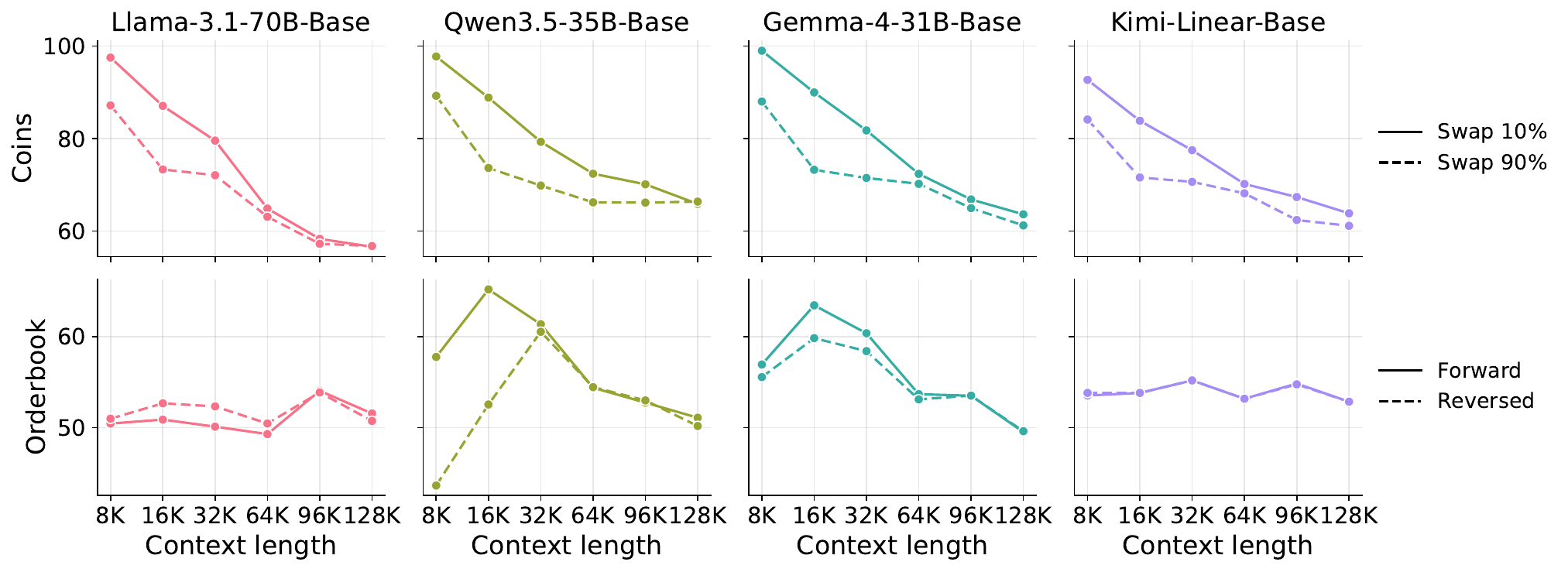}
    \caption{
        Difference in performance on variants of coins and orderbook. 
        For coins, we increase the probability of swapping as opposed to adding/removing coins from 10\% to 90\%. 
        For orderbook, we use reverse chronological order.
    }
    \label{fig:analysis_variants}
\end{figure}

We present the results for selected models in \Cref{fig:analysis_variants}.
We observe that there tends to be a significant gap between the two variants of Coins at shorter input lengths---increasing the percentage of swap operations led to as much as a 50\% decrease in performance across the four models. 
As the context length approaches 128K tokens, both variants converge to roughly the same performance around 60\%, which is worse than the naive baseline performance. 
This suggests that models are better at handling linear aggregation than other affine transformations in state tracking.

Furthermore, we see near-identical performance between the forward and reversed variants of Orderbook, with the exception of Qwen3.5-35B-Base at shorter contexts. 
The gap between the two variants is mostly under 3 absolute points across models, suggesting that models do not necessarily process state tracking tasks sequentially. 
Even Kimi-Linear, which leverages recurrent state modules, performs the same on the two variants \citep{kimiteam2025kimilinearexpressiveefficient}.

Our synthetic dataset construction enables other types of analysis as well, such as varying the sizing of the states and checking at what point the models fails. We leave these analysis for future work.

\subsection{Results on other benchmarks}

We evaluate subsets of HELMET and RULER, which consist of JSON KV from HELMET \citep{yen2025helmet}, NIAH MK 2, NIAH MK 3, and NIAH MV from RULER \citep{hsieh2024ruler}.
We show the average across these.
Additionally, we also evaluate other benchmarks, such as LongBench v2 \citep{bai-etal-2025-longbench}, MRCR \citep{vodrahalli2024michelangelolongcontextevaluations}, and GraphWalks \citep{openai2025gpt41}.
For MRCR, we evaluate the version with 4 needles released by OpenAI\footnote{\url{https://huggingface.co/datasets/openai/mrcr}}.
For GraphWalks, we evaluate the version released by the HELMET authors\footnote{\url{https://github.com/princeton-nlp/HELMET/tree/v2}} as the input lengths are measured by the Llama 3 tokenizer, and the maximum depth of the BFS variant is 5, and average the performance on the parent and the BFS variants.
For LongBench v2, we evaluate the long subset, but truncate all documents to 512K Llama 3 tokens so that the API models do not refuse due to exceeding the model context window.

\begin{figure}
    \centering
    \includegraphics[width=0.99\linewidth]{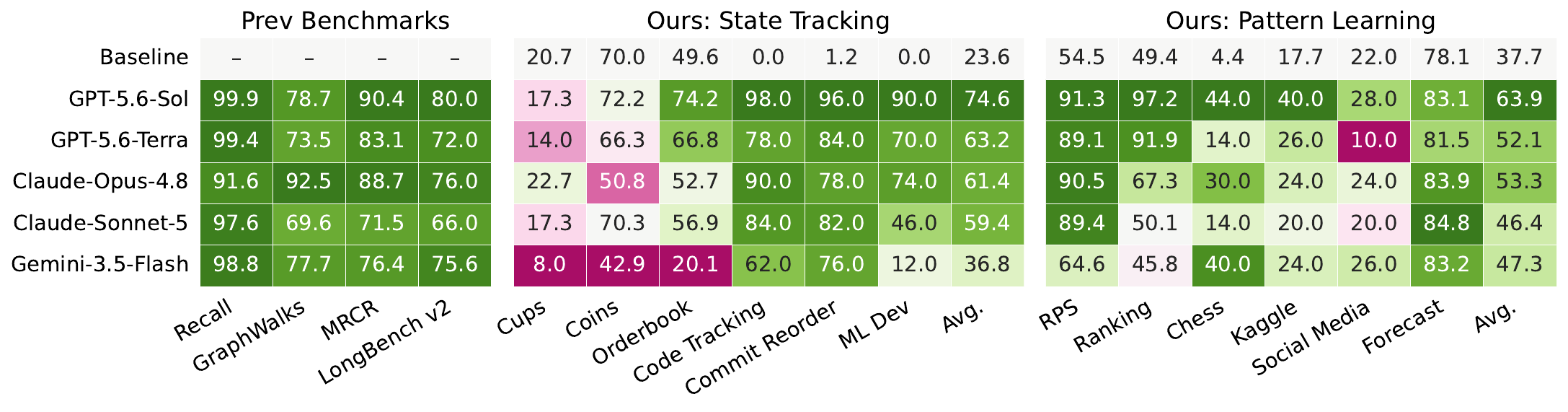}
    \caption{
        Results of API models at 512K input length, compared with previous synthetic benchmarks, recall (subsets of RULER and HELMET Recall), GraphWalks, MRCR, and LongBench v2.
        Recall is mostly saturated. 
        Despite the variation in performance, GraphWalks, MRCR, and LongBench v2 similarly only test for recall capabilities by design.
        \ours{} is conceptually distinct from previous benchmarks. 
    }
    \label{fig:results_512k_api_full}
\end{figure}

The full results for the API models at 512K input length are shown in \Cref{fig:results_512k_api_full}.
The full results for all models from input lengths 8K to 128K are shown in \Cref{tab:results_helmet}. LongBench v2 is excluded from this table because it does not support fine granularity of input lengths.

We evaluate the open-weight models on the same datasets as well, from input lengths 8K to 128K, and the results are shown in \Cref{tab:results_helmet}.
LongBench v2 is excluded because it does not have the different length buckets that we evaluated on.
We do not evaluate at 96K input length as many of the datasets did not generate data at that length. 

\begin{table}[t]
    \centering

\resizebox{\textwidth}{!}{%
\begin{tabular}{lrrrrrrrrrrrrrrrrrrrr}
\toprule
 & \multicolumn{5}{c}{Recall} & \multicolumn{5}{c}{GraphWalks} & \multicolumn{5}{c}{MRCR} & \multicolumn{5}{c}{Avg.} \\
\cmidrule(lr){2-6}\cmidrule(lr){7-11}\cmidrule(lr){12-16}\cmidrule(lr){17-21}
Model & 8K & 16K & 32K & 64K & 128K & 8K & 16K & 32K & 64K & 128K & 8K & 16K & 32K & 64K & 128K & 8K & 16K & 32K & 64K & 128K \\
\midrule
Llama-3.1-8B-Base & 99.7 & 99.6 & 98.3 & 88.7 & 54.0 & 6.4 & 2.3 & 1.6 & 1.1 & 2.5 & 0.0 & 0.0 & 0.0 & 0.0 & 0.0 & 58.8 & 57.6 & 56.6 & 51.0 & 31.6 \\
Llama-3.1-8B-Inst & 99.5 & 98.8 & 97.5 & 97.3 & 72.5 & 2.5 & 0.5 & 8.1 & 12.8 & 4.7 & 38.3 & 24.5 & 21.1 & 32.0 & 24.6 & 63.0 & 60.1 & 61.0 & 63.8 & 46.3 \\
Llama-3.1-70B-Base & \textbf{100.0} & 99.8 & 99.0 & 95.7 & 49.1 & 30.1 & 21.3 & 12.2 & 5.4 & 2.5 & 0.0 & 0.0 & 0.0 & 0.0 & 0.0 & 65.8 & 63.1 & 60.1 & 56.2 & 28.8 \\
Llama-3.1-70B-Inst & 99.9 & \textbf{99.9} & 99.1 & 97.9 & 37.8 & 24.2 & 47.9 & 42.3 & 33.6 & 7.5 & 57.2 & 33.9 & 28.1 & 27.9 & 12.0 & 72.2 & 75.6 & 72.7 & 69.6 & 25.5 \\
ProLong-8B-Base & 96.8 & 98.5 & 97.7 & 96.6 & 95.1 & 10.5 & 7.7 & 4.4 & 0.5 & 0.1 & 0.0 & 0.0 & 0.0 & 0.0 & 0.0 & 58.3 & 58.5 & 57.1 & 55.3 & 54.4 \\
ProLong-8B-Inst & 95.2 & 97.0 & 96.1 & 97.4 & 92.0 & 29.3 & 23.4 & 17.1 & 9.2 & 12.1 & 28.7 & 22.8 & 17.9 & 34.4 & 21.1 & 66.9 & 65.4 & 62.4 & 63.2 & 59.1 \\
GPT-OSS-120B & 88.2 & 83.1 & 86.5 & 83.5 & 52.6 & 75.8 & 67.1 & 61.2 & 44.8 & 24.1 & 75.1 & 54.4 & 43.1 & 31.7 & 19.9 & 82.8 & 74.4 & 73.1 & 65.0 & 39.8 \\
Qwen3-8B-Base & 84.7 & 80.5 & 48.7 & 39.1 & 0.0 & 11.4 & 4.6 & 1.7 & 0.2 & 29.0 & 0.0 & 0.0 & 0.0 & 0.0 & 0.0 & 51.7 & 47.3 & 28.3 & 22.4 & 8.3 \\
Qwen3-8B & 96.7 & 90.7 & 75.1 & 59.6 & 0.0 & 38.3 & 35.0 & 13.5 & 20.0 & 29.0 & 0.0 & 0.0 & 0.0 & 0.0 & 0.0 & 66.2 & 61.8 & 46.8 & 39.8 & 8.3 \\
Qwen3-30B-Base & 94.5 & 84.2 & 64.2 & 47.1 & 0.0 & 8.2 & 3.0 & 3.4 & 2.6 & 29.0 & 0.0 & 0.0 & 0.0 & 0.0 & 0.0 & 56.3 & 49.0 & 37.6 & 27.6 & 8.3 \\
Qwen3-30B & 98.3 & 90.6 & 81.9 & 63.1 & 0.0 & 59.1 & 30.1 & 29.3 & 16.5 & 29.0 & 0.0 & 0.0 & 0.0 & 0.0 & 0.0 & 73.1 & 60.4 & 55.2 & 40.7 & 8.3 \\
Qwen3-30B-Think-2507 & \textbf{100.0} & 99.8 & 99.9 & 97.8 & 98.3 & 74.7 & 50.1 & 43.9 & 32.4 & 26.1 & 0.0 & 0.0 & 0.0 & 0.0 & 0.0 & 78.5 & 71.3 & 69.6 & 65.1 & 63.6 \\
Qwen3-30B-Inst-2507 & 99.8 & 99.7 & 99.6 & 98.2 & 98.5 & 33.4 & 27.6 & 23.4 & 21.9 & 14.0 & 30.4 & 24.5 & 25.4 & 35.0 & 26.7 & 70.9 & 68.3 & 67.2 & 67.4 & 64.1 \\
Qwen3.5-9B-Base & 98.1 & 98.0 & 97.0 & 97.6 & 96.4 & 28.0 & 15.4 & 12.7 & 13.8 & 8.6 & 0.0 & 0.0 & 0.0 & 0.0 & 0.0 & 64.1 & 60.4 & 59.0 & 59.7 & 57.6 \\
Qwen3.5-9B & 95.5 & 94.8 & 95.6 & 96.1 & 92.2 & 77.7 & 74.4 & 75.7 & 75.1 & 73.0 & 0.0 & 0.0 & 0.0 & 0.0 & 0.0 & 76.8 & 75.4 & 76.3 & 76.4 & 73.6 \\
Qwen3.5-35B-Base & 95.9 & 96.7 & 97.9 & 98.0 & 98.1 & 28.8 & 23.6 & 29.8 & 29.5 & 27.9 & 0.0 & 0.0 & 0.0 & 0.0 & 0.0 & 63.0 & 62.0 & 64.5 & 64.4 & 64.0 \\
Qwen3.5-35B & 98.2 & 97.4 & 97.2 & 95.3 & 93.8 & \textbf{78.3} & 74.7 & 74.9 & \textbf{75.9} & \textbf{78.3} & 0.0 & 0.0 & 0.0 & 0.0 & 0.0 & 78.5 & 77.0 & 76.9 & 76.1 & 75.9 \\
Gemma-4-12B-Base & 98.9 & 99.1 & 98.6 & 98.4 & 95.4 & 14.9 & 14.5 & 11.6 & 2.5 & 1.0 & 0.0 & 0.0 & 0.0 & 0.0 & 0.0 & 60.8 & 60.7 & 59.6 & 56.9 & 54.8 \\
Gemma-4-12B-Inst & 99.7 & 99.6 & 99.4 & 98.8 & 94.5 & 42.9 & 43.5 & 53.8 & 45.2 & 33.2 & 73.3 & 53.0 & 48.6 & 38.9 & 29.9 & 79.7 & 76.9 & 79.1 & 74.9 & 67.7 \\
Gemma-4-31B-Base & 99.8 & 99.8 & 99.7 & 99.4 & 98.8 & 37.7 & 34.5 & 28.5 & 21.8 & 9.7 & 0.0 & 0.0 & 0.0 & 0.0 & 0.0 & 67.8 & 66.9 & 65.1 & 63.1 & 59.3 \\
Gemma-4-31B-Inst & \textbf{100.0} & \textbf{99.9} & \textbf{99.9} & \textbf{99.6} & \textbf{99.1} & 75.1 & \textbf{77.1} & \textbf{76.7} & 70.0 & 42.7 & 91.1 & 73.8 & 71.7 & 64.7 & 62.8 & \textbf{91.6} & \textbf{89.7} & \textbf{89.3} & \textbf{86.1} & 77.8 \\
Nemotron-3-30B-Base & 97.1 & 92.9 & 91.9 & 81.7 & 77.0 & 6.5 & 2.2 & 1.2 & 0.3 & 0.4 & 0.0 & 0.0 & 0.0 & 0.0 & 0.0 & 57.4 & 53.7 & 52.9 & 46.8 & 44.1 \\
Nemotron-3-30B & 88.8 & 83.6 & 67.6 & 65.8 & 52.0 & 49.4 & 30.5 & 28.7 & 26.5 & 27.8 & 0.0 & 0.0 & 0.0 & 0.0 & 0.0 & 64.8 & 56.5 & 46.8 & 45.2 & 37.7 \\
Kimi-Linear-Base & 98.6 & 99.1 & 98.9 & 99.2 & 97.7 & 16.9 & 11.4 & 8.5 & 4.9 & 4.3 & 0.0 & 0.0 & 0.0 & 0.0 & 0.0 & 61.2 & 59.9 & 59.0 & 58.1 & 57.1 \\
Kimi-Linear-Inst & 99.2 & 98.8 & 98.9 & 98.9 & 98.4 & 68.5 & 66.3 & 61.2 & 58.7 & 55.1 & \textbf{94.3} & \textbf{98.7} & \textbf{91.8} & \textbf{87.6} & \textbf{81.5} & 89.8 & 89.5 & 87.1 & 85.8 & \textbf{83.6} \\
\bottomrule
\end{tabular}%
}

    \caption{Results of all models on previous benchmarks: Recall is a mix of RULER and HELMET Recall, GraphWalks, MRCR, and the average across these categories. LongBench v2 is excluded as it does not support fine granularity of input lengths.}
    \label{tab:results_helmet}
\end{table}

\subsection{Additional results}
\label{app:more-results}

\begin{figure}
    \centering
    \includegraphics[width=0.99\linewidth]{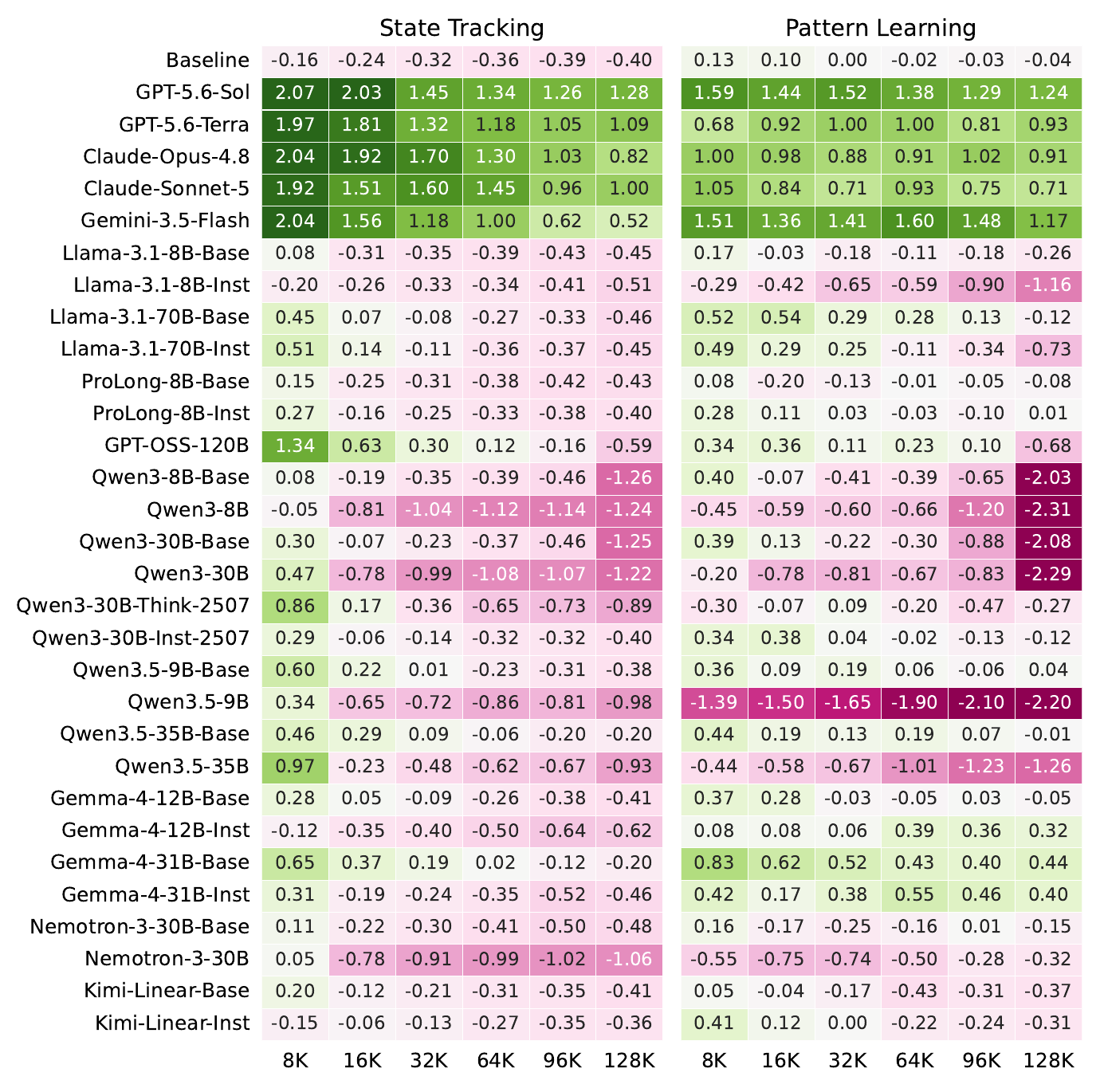}
    \caption{
        Results in z-scores averaged across each category from 8K to 128K tokens.
        We first calculate the mean and variance for a dataset across 30 evaluated models and 6 lengths and normalize all scores into the z-score to reduce bias for any particular dataset in the category average.
        Absolute scores and all models are shown in \Cref{fig:results_128k_full}.
    }
    \label{fig:results_128k_avg}
\end{figure}

\begin{figure}
    \centering
    \includegraphics[width=0.99\linewidth]{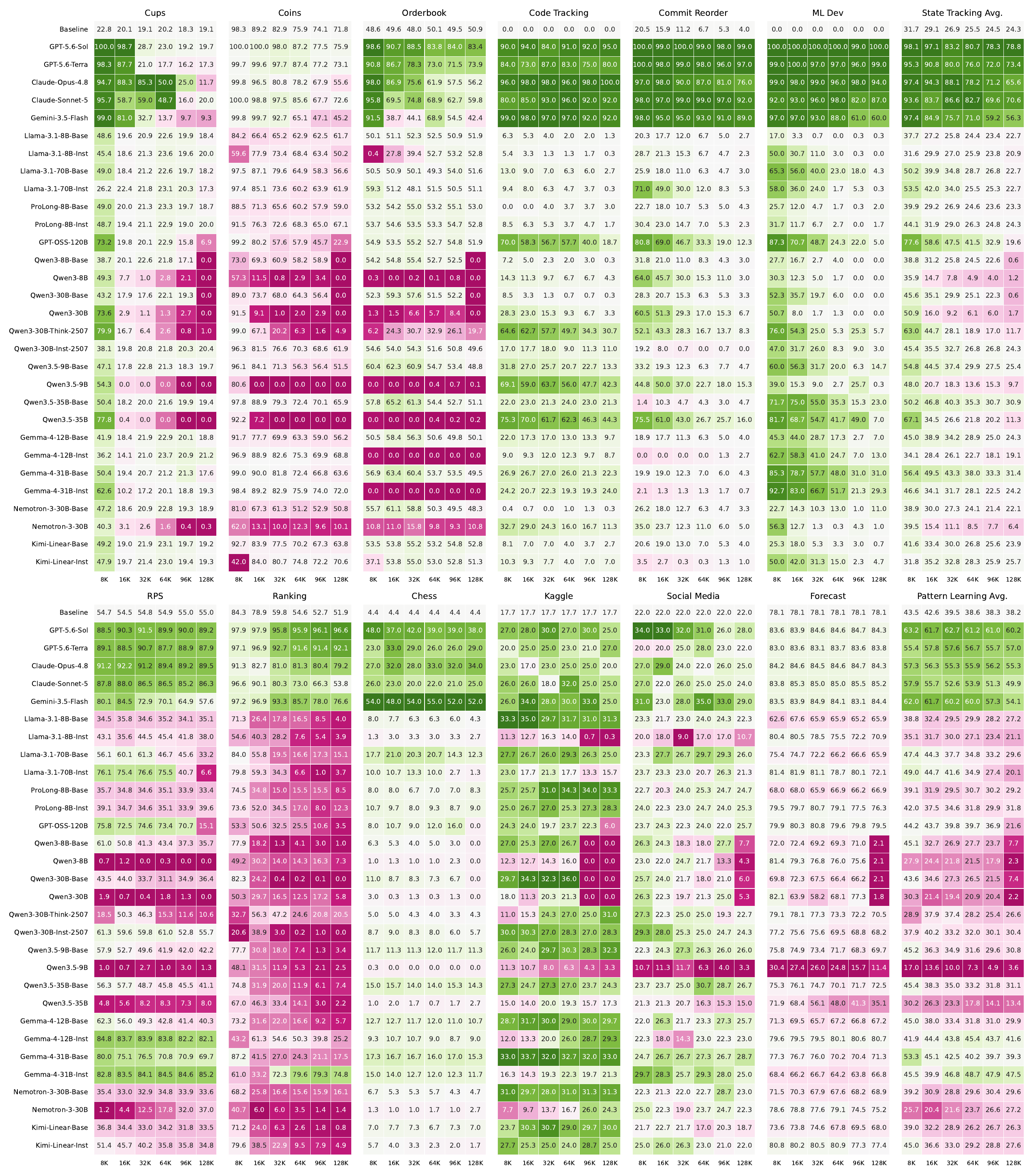}
    \caption{
        Main results across six datasets for each category and input lengths from 8K to 128K tokens for all 30 evaluated models.
    }
    \label{fig:results_128k_full}
\end{figure}

\begin{figure}[t!]
    \centering
    \includegraphics[width=0.99\linewidth]{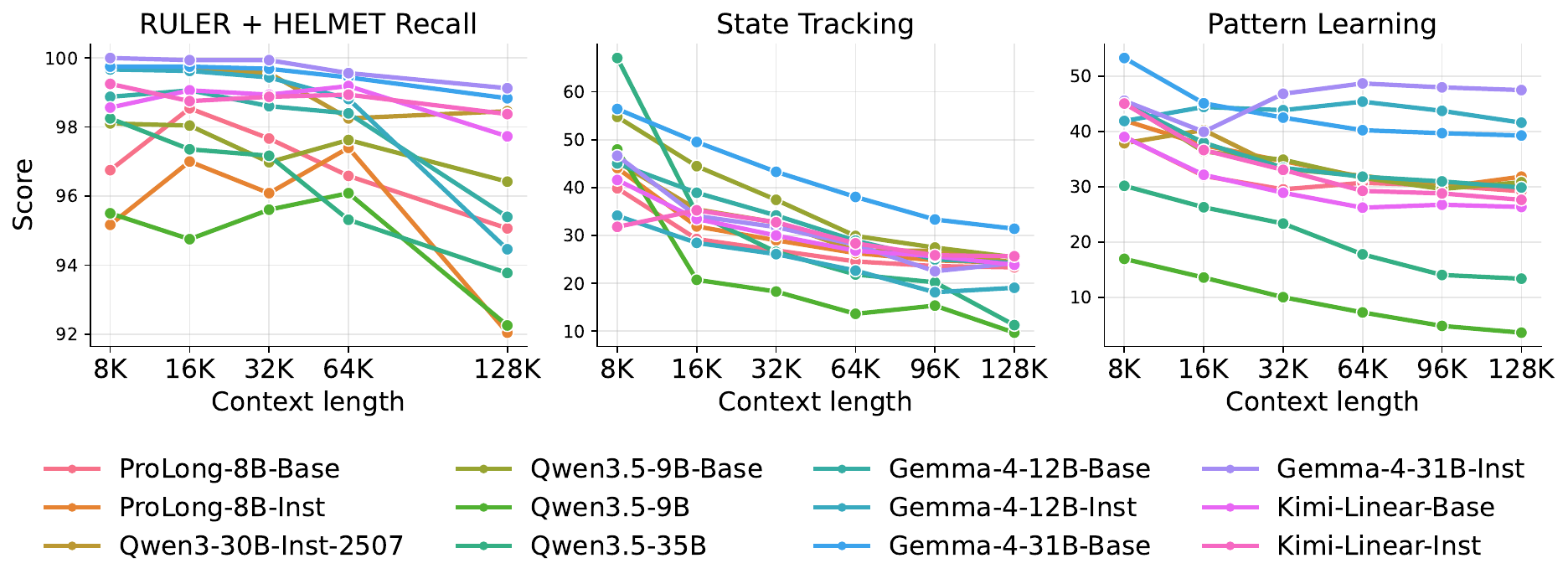}
    \caption{
        Average absolute performance on each capability.
        Recall tasks are saturated, and models observe relatively limited degradation, whereas \ours{} tests for new capabilities that are still not saturated and experience more degradation.
    }
    \label{fig:results_128k_comp_abs}
\end{figure}

\begin{figure}
    \centering
    \includegraphics[width=0.99\linewidth]{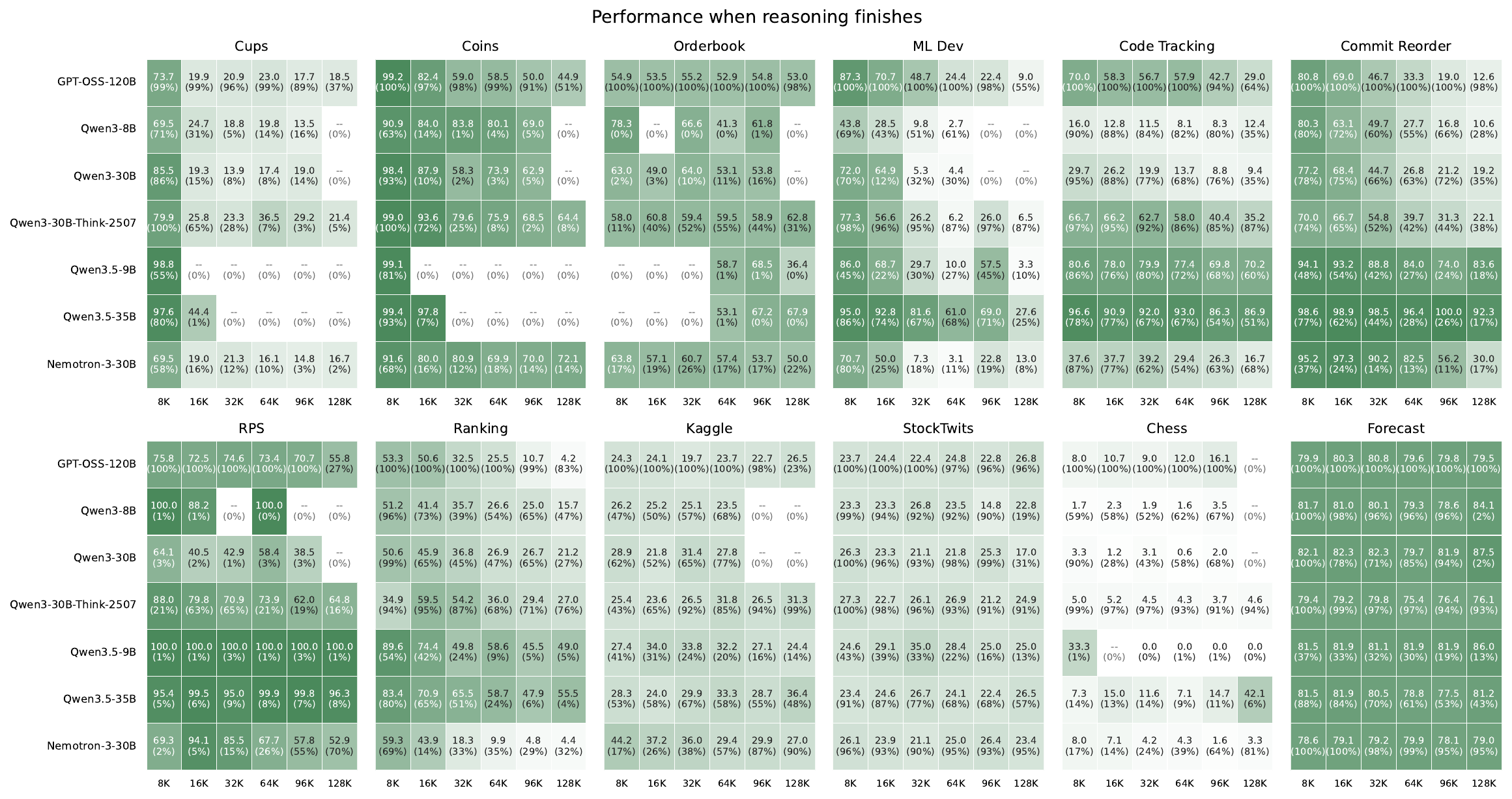}
    \caption{
        Results of thinking models when thinking finishes. The default generation budget is up to 16K tokens.
    }
    \label{fig:analysis_thinking_finish_full}
\end{figure}

In this section, we show additional results: the results in z-scores in \Cref{fig:results_128k_avg}, the full results in \Cref{fig:results_128k_full}, the absolute performance on each capability in \Cref{fig:results_128k_comp_abs}, and results of thinking models if they were to finish in \Cref{fig:analysis_thinking_finish_full}.

\begin{table}[]
    \centering

        \resizebox{0.46\textwidth}{!}{%
        \begin{tabular}{lrr}
        \toprule
                   dataset &     mean &      std \\
        \midrule
                     Chess & 0.118778 & 0.122106 \\
             Code Tracking & 0.297739 & 0.324307 \\
                     Coins & 0.623230 & 0.301486 \\
                Coins Swap & 0.543219 & 0.283638 \\
            Commit Reorder & 0.293655 & 0.334649 \\
                      Cups & 0.257087 & 0.227134 \\
                  Forecast & 0.710904 & 0.160736 \\
                    Kaggle & 0.228704 & 0.088361 \\
                    ML Dev & 0.358389 & 0.344583 \\
                 Orderbook & 0.428420 & 0.266986 \\
        Orderbook Reversed & 0.366124 & 0.236013 \\
                       RPS & 0.478728 & 0.283423 \\
                   Ranking & 0.371046 & 0.321393 \\
              Social Media & 0.229648 & 0.055132 \\
        \bottomrule
        \end{tabular}
        }
        
    \caption{Means and standard deviations across all models on input lengths from 8K to 128K.}
    \label{tab:results_mean_std}
    
\end{table}

\end{document}